\documentclass[11pt]{article}

\usepackage[preprint]{acl}

\usepackage{times}
\usepackage{latexsym}
\DeclareUnicodeCharacter{2206}{\ensuremath{\Delta}}

\usepackage[T1]{fontenc}
\usepackage[utf8]{inputenc}
\usepackage{amsmath}
 
\usepackage{microtype}

\usepackage{inconsolata}

\usepackage{graphicx}
\usepackage{times}
\usepackage{adjustbox}    % 调整/缩放盒子

\usepackage{booktabs}   % For professional-looking horizontal lines
\usepackage{tabularx}   % For auto-adjusting column widths
\usepackage{amsmath}    % Required for \substack and \text commands
\usepackage{amsmath}        % 用于数学公式
\usepackage{ragged2e}       % 用于改善长文本列的对qi
\usepackage{threeparttable} % 用于在表格下方添加注释
\usepackage{tcolorbox}
\tcbuselibrary{breakable} %breakable：支持跨页
\usepackage{caption}     % 用于表格标题
\usepackage[table]{xcolor} % 用于给表格单元格上色
\usepackage{siunitx}     % 用于对齐数字、处理单位
\usepackage{makecell}    % 用于在单元格内换行

\usepackage{colortbl}      % \rowcolor、\cellcolor
\usepackage{threeparttable}% 表格脚注环境
\usepackage{subcaption}    % \subcaption* 用于子表标题
\usepackage[table,xcdraw]{xcolor} % 颜色（含表格专用可选项）
\usepackage{fancyvrb}
\usepackage{listings}
\usepackage{xcolor}
\usepackage{float}
\usepackage{algorithm}
\usepackage{algorithmic}
\usepackage{graphicx}
\usepackage{amssymb}
\usepackage{placeins}  % Ensure figures stay within their section or paragraph
\definecolor{highDeep}{HTML}{4A90E2}   % 最大值
\definecolor{highLite}{HTML}{D4E4F9}   % 次大值
\definecolor{lowDeep} {HTML}{F28E8E}   % 最小值
\definecolor{lowLite} {HTML}{F8DADA}   % 次小值
\definecolor{myblue}{rgb}{0.15, 0.15, 0.5}
\definecolor{myblue}{rgb}{0.15, 0.15, 0.5}
\definecolor{mylightpurple}{rgb}{0.8, 0.7, 0.9}
\usepackage[utf8]{inputenc}
\usepackage{listings} 

\usepackage{booktabs} % 用于 toprule, midrule, bottomrule
\usepackage{ragged2e} % 用于 RaggedRight

\definecolor{myblue1}{rgb}{0.2, 0.2, 0.8}
\definecolor{mygray1}{rgb}{0.9, 0.9, 0.9}

\usepackage[utf8]{inputenc}

\usepackage{colortbl}   % \rowcolor
\renewcommand{\arraystretch}{1.25}
\newcolumntype{C}[1]{>{\centering\arraybackslash}m{#1}}

\usepackage{minted}     % 需 -shell-escape
\usepackage[normalem]{ulem} % normalem 避免把 \emph 也改成带线
\definecolor{highlightcolor}{gray}{0.9}
\usepackage{listings}
\definecolor{myred}{RGB}{178, 34, 34} % 砖红色
\definecolor{mygreen}{RGB}{34,139,34}   % 林绿（常见标准色）
\definecolor{myred2}{RGB}{237, 211, 210} % 自定义一种浅红色
\definecolor{mygreen2}{RGB}{198, 232, 206} % 自定义一种浅绿色
\definecolor{myblue2}{RGB}{218,232,252}

\definecolor{codegreen}{rgb}{0,0.6,0}
\definecolor{codegray}{rgb}{0.5,0.5,0.5}
\definecolor{codepink}{RGB}{252, 142, 172}
\definecolor{codepurple}{rgb}{0.58,0,0.82}
\definecolor{backcolour}{RGB}{245,245,245}

\lstdefinestyle{mystyle}{
    backgroundcolor=\color{backcolour},   
    commentstyle=\color{magenta},
    keywordstyle=\color{blue},
    numberstyle=\tiny\color{codegray},
    stringstyle=\color{codepurple},
    basicstyle=\fontfamily{\ttdefault}\footnotesize,
    breakatwhitespace=false,         
    breaklines=true,                 
    keepspaces=true,    
    frame=single,
    numbersep=5pt,                  
    showspaces=false,                
    showstringspaces=false,
    showtabs=false,                  
    tabsize=2,
    classoffset=1, % starting new class
    keywordstyle=\color{violet},
    classoffset=0,
}
\lstdefinelanguage{JavaScript}{
  keywords={typeof, new, true, false, catch, function, return, null, catch, switch, var, if, in, while, do, else, case, break},
  keywordstyle=\color{blue}\bfseries,
  ndkeywords={class, export, boolean, throw, implements, import, this},
  ndkeywordstyle=\color{darkgray}\bfseries,
  identifierstyle=\color{black},
  sensitive=false,
  comment=[l]{//},
  morecomment=[s]{/*}{*/},
  commentstyle=\color{purple}\ttfamily,
  stringstyle=\color{red}\ttfamily,
  morestring=[b]',
  morestring=[b]"
}
\usepackage{hyperref}
\usepackage{url}
\title{\textit{DataGovBench}: Benchmarking LLM Agents for Real-World Data Governance Workflows}

\author{
  Zhou Liu\textsuperscript{1}\thanks{Equal contribution.} \quad
  Zhaoyang Han\textsuperscript{1}\footnotemark[1] \quad
  Guochen Yan\textsuperscript{1} \quad 
  Hao Liang\textsuperscript{1} \quad \\
  Bohan Zeng\textsuperscript{1} \quad
  Xing Chen\textsuperscript{2} \quad
  Yuanfeng Song \textsuperscript{2}  \thanks{Corresponding authors.} \quad 
  Wentao Zhang\textsuperscript{1} \footnotemark[2] \\
  \textsuperscript{1}Peking University, China\\
  \textsuperscript{2}ByteDance, China
}

\begin{document}
\maketitle
\begin{abstract}
Data governance ensures data quality, security, and compliance through policies and standards—a critical foundation for scaling modern AI development. Recently, Large Language Models (LLMs) have emerged as a promising solution for automating data governance by translating user intent into executable transformation code. However, existing benchmarks for automated data science often emphasize snippet-level coding or high-level analytics, failing to capture the unique challenge of data governance: ensuring the correctness and quality of the data itself. To bridge this gap, we introduce DataGovbench, a benchmark featuring 150 diverse tasks grounded in real-world scenarios, built on data from actual cases. DataGovbench employs a novel "reversed-objective" methodology to synthesize realistic noise and utilizes rigorous metrics to assess end-to-end pipeline reliability. Our analysis on DataGovbench reveals that current models struggle with complex, multi-step workflows and lack robust error-correction mechanisms. Consequently, we propose DataGovAgent, a framework utilizing a Planner-Executor-Evaluator architecture that integrates constraint-based planning, retrieval-augmented generation, and sandboxed feedback-driven debugging. Experimental results show that DataGovAgent significantly boosts the Average Task Score (ATS) on complex tasks from 39.7 to 54.9 and reduces debugging iterations by over 77.9\% compared to general-purpose baselines.

\end{abstract}

\section{Introduction}

Data fuels analytics and machine intelligence, yet the work required to make data trustworthy remains manual. Studies report \citep{ahmadi2024performance} that practitioners spend the majority of their time cleaning, standardizing, integrating, and validating data rather than modeling it, turning skilled analysts into ``data janitors” and creating a persistent bottleneck in the data value chain \citep{hosseinzadeh2023data}. Code-centric Extract, Transform, Load (ETL) pipelines and handwritten SQL/Python are powerful but inefficient in the face of evolving data structures and data heterogeneity \citep{yang2025etl,dinesh2024etl}, costly to maintain, and slow to adapt to evolving business rules.

LLMs promise an alternative: specify governance intent in natural language and generate the required transformation code automatically \citep{pahune2025governance,park2025dataverse}. However, progress is hampered by the lack of adequate evaluation standards. Existing benchmarks for automated data science often emphasize snippet-level coding or high-level analytics like summaries and reports, failing to capture the unique challenges of data governance, \textit{which is the correctness and quality of data itself.} 

% Current evaluation methodologies are overly coarse-grained. In the realm of data governance, executability does not equate to data trustworthiness. To truly entrust data pipelines to LLMs, we must move beyond simple syntax checks and evaluate the correctness of the entire operator graph (DAG). 

To address this limitation, we introduce DataGovbench, a hierarchically designed benchmark for natural-language-driven data governance. It contains 150 real-world tasks (100 operator-level; 50 DAG-level) covering six scenarios: Filtering, Refinement, Imputation, Deduplication \& Consistency, Data Integration, and Classification \& Labeling. DataGovbench's key innovations include: i) an exceptionally comprehensive range of tasks, covering both atomic and multi-step processes, tasks on structured and unstructured data, as well as six distinct tasks grounded in real-world production environments. ii) a novel \textit{“reversed‑objective"} methodology—that inverts the original task goal to programmatically generate task‑specific noise—to synthesize realistic and measurable noise; and iii) auto-generated, task-specific evaluation scripts that provide accurate normalized scores and standardized metrics to compare ground truth dataset with processed dataset—Code Runnable Rate (CRR), Task Success Rate (TSR), and Average Task Score (ATS)—ensuring a principled and reproducible assessment.

% 2) a longest‑common‑subsequence–aware (LCS-aware) sequencing algorithm that constructs compositionally deep DAG tasks with minimal pairwise overlap;

However, a robust benchmark is only half of the solution. When evaluated on DataGovbench, we find that even SOTA LLMs \citep{openai2025gpt5,deepseekv3,openai2024gpt4o} baselines and general-purpose agent frameworks \citep{qian-etal-2024-chatdev,li2023camel} exhibit a significant performance gap. They struggle to decompose complex instructions, generate logically correct multi-step pipelines, and debug errors, resulting in low task success rates. This indicates that, for data governance workflows, their architectures are not yet equipped with robust planning, strong grounding in established practices, or systematic debugging mechanisms.

To bridge this performance gap, we propose DataGovAgent, an end-to-end framework designed to translate natural language into executable governance workflows. The system operates through a sequential multi-agent pipeline (Agentic Assembly Line) consisting of three specialized roles: i) a Planner that converts user intent into a high-level DAG of data operations and adds runtime checks to detect errors during execution; ii) an Executor that uses retrieval-augmented generation over a curated library to minimize generation errors and improve code quality; and iii) an Evaluator that manages an iterative debugging loop in a sandbox, utilizing execution feedback to refine the code until it is both runnable and functionally correct.

On DataGovbench-150, DataGovAgent significantly outperforms strong baselines. With GPT-5, it raises the Task Success Rate (TSR)—the core metric representing the proportion of tasks where business objectives are fully achieved—to 64\% on operator-level tasks (+15 pp) and 60\% on DAG-level tasks (+14 pp). Furthermore, compared to the strongest baseline ChatDev, DataGovAgent demonstrates superior efficiency by drastically reducing Average Debug Iterations (ADI) from 14.89 to 3.29.

In summary, our contributions are twofold:
\begin{itemize}
    \item We introduce DataGovbench, the first hierarchical-designed benchmark for data governance automation, which features 150 realistic tasks based on real-world sources, noisy data synthesis and a rigorous, multi-metric evaluation protocol to address the critical gap in assessing end-to-end pipeline correctness.
    \item We propose DataGovAgent, a framework that significantly improves task success by translating natural language into verified governance pipelines through the integration of contract-guided planning, retrieval-augmented code generation, and feedback-driven debugging.
\end{itemize}

\section{Related Works}

%\subsection
\noindent \textbf{Data Science Benchmarks and LLM Evaluation.} The rapid evolution of LLMs has catalyzed comprehensive evaluation frameworks for automated data science capabilities. Early benchmarks like DS-1000~\citep{lai2023ds1000} focused on snippet-level code generation for data science libraries, extended by DA-Code~\citep{huang2024dacode} for task-level evaluation in interactive environments. GAIA ~\cite{mialon2023gaiabenchmarkgeneralai} aims at assessing model's ability  of handling real-world tasks including tabular data analysis through coding. However, it serves as a general testbed for LLMs and the data science related tasks tend to be overly simple due to question design and data size. Recently, DataSciBench \citep{zhang2025datascibench}, which provides systematic LLM agent evaluation with 25 multidimensional metrics across complete data science workflows, and ScienceAgentBench \citep{chen2025scienceagentbench}, which targets rigorous assessment for data-driven scientific discovery, have been proposed (see Appendix \ref{appendix:related works table} for detailed benchmark comparison). Unfortunately, all existing benchmarks rarely focus on the model's ability to improve data quality and enhance its usability and trustworthiness.

Contemporary LLM evaluation has shifted toward sophisticated multidimensional assessment. HumanEval Pro \citep{yu2025humanevalpro} introduces self-invoking code generation requiring progressive reasoning capabilities, while mHumanEval \citep{raihan2025mhumaneval} extends multilingual code evaluation. LiveBench \citep{white2025livebench} addresses contamination issues in LLM evaluation with challenging, dynamic benchmarks (see Appendix A). These frameworks demonstrate significant performance variations, with SOTA models achieving 96.2\% on HumanEval but declining to 76.2\% on complex tasks.

\noindent \textbf{Data Science Agents and Automation.} Data science agents have evolved from simple code generators to comprehensive autonomous systems. Data Interpreter \citep{hong2024datainterpreter} employs hierarchical graph modeling for dynamic problem decomposition, while recent developments include AutoMind \citep{ou2025automind}, offering adaptive knowledgeable agents for automated data science, and AutoML-Agent~\citep{trirat2025automlagent}, providing multi-agent frameworks for full-pipeline AutoML.

Current research emphasizes end-to-end workflow automation with minimal human intervention~\citep{sun2024survey}. TheAgentCompany \citep{xu2025theagentcompany} benchmarks LLM agents on consequential real-world tasks, while comprehensive surveys \citep{baek2025automation,wang2025survey} highlight the transition from automation to autonomy in scientific discovery. These systems integrate planning, reasoning, reflection, and multi-agent collaboration capabilities. However, specialized data governance benchmarks remain limited. This gap highlights the necessity for benchmarks like our proposed \textbf{DataGovbench}.

Our work contributes through efficient data annotation pipelines generating customized evaluation scripts with standardized metrics including Code Runnable Rate (CRR), Task Success Rate (TSR), and Average Task Score (ATS), addressing gaps in governance-focused evaluation methodologies.

\section{\textit{DataGovbench:} A New Benchmark for Data Governance Automation}

\label{sec:method}
\begin{figure}[t!]
    \centering
    \includegraphics[width=0.5\textwidth]{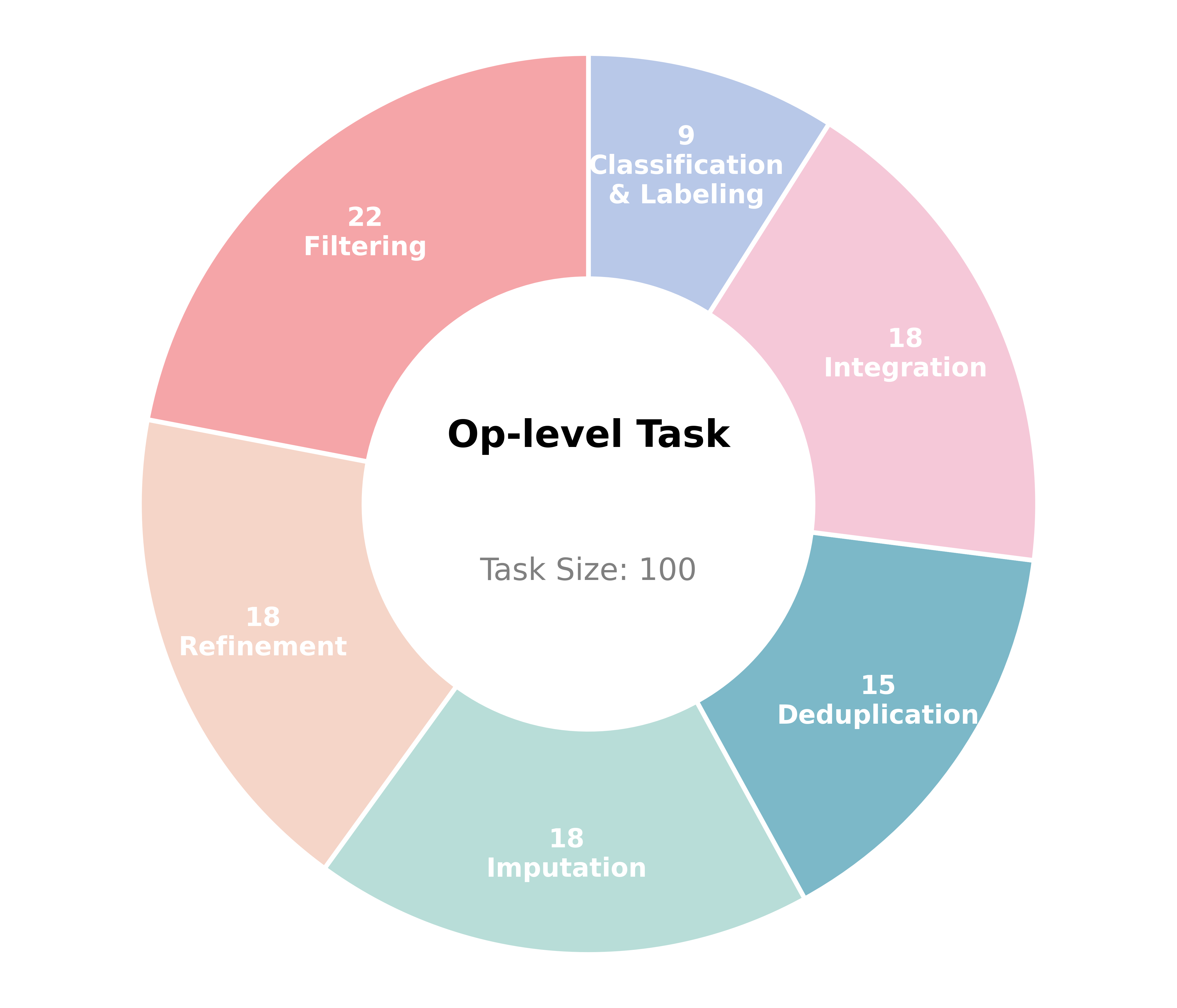}
    \caption{Distribution of the 100 operator-level tasks in DataGovbench across six governance categories: Filtering (22), Refinement (18), Imputation (18), Deduplication \& Consistency (15), Integration (18), and Classification \& Labeling (9). The split balances coverage of common governance operations while slightly emphasizing filtering and transformation tasks.}
    \label{fig:task distribution}
\end{figure}

DataGovbench is a hierarchically designed data science benchmark dedicated to evaluating models' capabilities in performing data governance tasks. It comprises 150 real-world data governance problems, including 100 operator-level tasks and 50 DAG-level tasks. DataGovbench comprehensively covers common scenarios encountered in real-life data governance workflows, including Filtering, Refinement, Imputation, Deduplication \& Consistency, Data Integration, and Classification \& Labeling. For each carefully curated NL task description, we proposed a reliable pipeline to synthesize ground-truth data and noisy data, accompanied by customized evaluation scripts to ensure precise and normalized scoring. 

\begin{figure*}[t!]
    \centering
    \includegraphics[width=\textwidth]{figures/gen_gemini2.png}
    \caption{Illustration of the semi-automated pipeline designed for building DataGovbench, including real-world source data curation, hierarchical task objective design, noisy data synthesis, task-specific evaluation and DAG-level task construction.}
    \label{fig:benchmark}
    \vspace{-15pt}
\end{figure*}

\textbf{Overview of Benchmark Creation.} To construct a hierarchical and realistic evaluation set for LLM-based data governance agents, we design a semi-automated pipeline comprising five stages: (1) data collection and column selection, (2) task objective definition, (3) noisy data synthesis, (4) generation of task-specific evaluation scripts and (5) DAG-level Tasks Construction(see Figure~\ref{fig:benchmark}; details in Sections~\ref{section:3.1}--\ref{section:3.5}). Statistics and examples are illustrated in Figure~\ref{fig:task distribution}.

\subsection{Real-world Data Source}
\label{section:3.1}
To ensure comprehensive coverage of real-world scenarios, we curated 30 tables sourced from~\citet{statista_2025}, spanning diverse domains such as tourism, eco-commerce, sports, and others. We retained only task-relevant columns (e.g., the \texttt{date} field for format normalization tasks) and necessary confounding columns (such as \texttt{birth\_date}, which agents are not required to modify), thus maintaining data integrity and minimizing unnecessary noise. Moreover, the columns selected include both structured and unstructured contents, intended for more diverse task design. These carefully selected and preprocessed datasets serve as the basis for synthesizing task descriptions, as detailed in Section~\ref{section:3.2}.

\subsection{Hierarchical Task Objective Design}
\label{section:3.2}
DataGovbench comprises 100 Operator-level tasks and 50 DAG-level tasks. For Operator-level tasks, we designed six scenarios commonly encountered in real-world data governance, including Filtering, Refinement, Imputation, Deduplication \& Consistency, Data Integration, and Classification \& Labeling. We recruited five postgraduate students majoring in Computer Science, each with significant experience in data analysis and data science, to annotate Operator-level task descriptions. We provided a clear definition of the task description schema in advance and conducted a thorough manual validation to ensure both accuracy and clarity. As a result, 32\% of the tasks were removed due to lack of feasibility, for instance, tasks with unclear objectives or ambiguous data requirements, and a final set of 100 task descriptions was retained. The distribution of tasks in these scenarios is illustrated in \textbf{Figure \ref{fig:task distribution}}.

\subsection{Noisy data synthesis}
\label{section:3.3}
To ensure robust and accurate evaluation of data governance models, it's essential to simulate real-world data disruptions in a controlled manner. This process is divided into two distinct steps, ensuring that only relevant noise is added while simulating real-world data environment. This approach guarantees fair scoring by preventing unnecessary noise from influencing the results.

\textbf{Generate a Reversed Task Objective.} The first step involves generating a reversed task objective based on the provided data examples and the original task objective. This reversed objective shifts the focus from achieving the task goal (e.g., classification, imputation) to deliberately introducing noise into the data.  For example, if the original task involves classifying data, the reversed task objective will focus on how to introduce noise such as mislabeling or irrelevant features. See the prompt template in \textbf{Prompt \ref{prompt:reverse}}.

\textbf{Generate Code to Introduce Noise.} In the second step, the model uses the reversed task objective, along with the provided data examples, to generate the actual code that will introduce the noise into the data. This code will implement the instructions described in the reversed objective—whether that involves adding missing values, creating duplicates, or generating irrelevant features. The goal is to transform the data in a way that makes it "disrupted", allowing the model to be tested with noisy inputs. See the prompt template in \textbf{Prompt \ref{prompt:noise}}.

At last, we manually check every data file, ensuring no extra noise is introduced because of model hallucination. During this process, 12 Operator-level tasks failed to pass the check. So we manually modified the data manipulation scripts and re-executed them. This two-step approach allows for a targeted and methodical introduction of noise, ensuring that the noise is task-specific and realistic, which helps in robustly evaluating the model's performance.

\subsection{Task-Specific Evaluation}
\label{section:3.4}

Unlike existing benchmarks, which typically use numbers, strings, or similar forms as ground truth, DataGovbench uses original data sheets as its ground truth. Given the diverse nature of task categories and the varying characteristics of the source data, customized evaluation is essential for each task. Specifically, we designed a prompt template (See \textbf{Prompt \ref{prompt:eval}} to generate the task-specific evaluation scripts.
Each task's script compares the original data with the processed data and outputs a quantitative score between 0 and 1, reflecting the model’s effectiveness in completing the task. Evaluation logic is adjusted based on the specific nature of the task to ensure a precise assessment; a detailed breakdown for each Operator-level task category is provided in Table~\ref{tab:op_eval_metrics}.
% in the \textbf{Appendix \ref{appendix:Tasks Eval}} .

\textbf{Consistency Check}  After preparing the evaluation scripts, we run them on both the ground truth data and the input data. The ground truth should yield a score of 1.0, while the raw data should score below 0.3. If these conditions are not met, we manually adjust either the raw data or the scripts to ensure compliance with the standard.

\subsection{DAG-level Tasks Construction}
\label{section:3.5}

To construct DAG-level tasks, we first rank the operator-level tasks by averaging the scores of GPT-5 and DeepSeek-V3 \citep{deepseekv3}, thereby mitigating the bias introduced by relying solely on a single closed-source model or an open-source model. We then select the 50 worst-performing Operator-level tasks as seed cases, treating the remaining tasks as candidates. Annotators (see details in \ref{section:3.2}) are tasked with creating a set of Op-level task sequences following these rules: (1) the sequence length should be longer than 2; (2) try not to repeat same subsequences; and (3) the operation sequences must be logical. For example, transforming a date into two different formats consecutively would be nonsensical. Finally, we use the candidate tasks to modify the created sequences if needed, reducing similarity as much as possible. Given these sequences, we employ the prompt template provided in the \textbf{Prompt \ref{prompt:dag}} to construct new natural language task objectives.We then synthesize the noisy data for these tasks using the method mentioned in Section \ref{section:3.3}.

To evaluate models on these tasks, we reuse the evaluation scripts for existing tasks and the final score is calculated based on the weighted average of scores from the Operator-level tasks. We still use the average scores of GPT-5 and DeepSeek-V3 \citep{deepseekv3} to mitigate the bias of any single source. The weights are determined by the following formula:
\begin{equation}
w_i = \frac{1}{1 + \alpha \cdot \text{score}_i}  
\end{equation}
where \( w_i \) is the weight of task \( i \), \( \alpha \) is a parameter that adjusts the influence of lower task scores, and \( \text{score}_i \) is the average performance score of two solutions for each individual task. So finally, the DAG-level task score is calculated by:
\begin{equation}
\text{DAG-level score} = \sum_{i=1}^{n} w_i \cdot \text{score}_i
\end{equation}

\section{\textit{DataGovAgent:} An End-to-End NL2GovDAG Framework for Data Governance}

\begin{figure*}[t]
    \centering
    \includegraphics[width=1\textwidth]{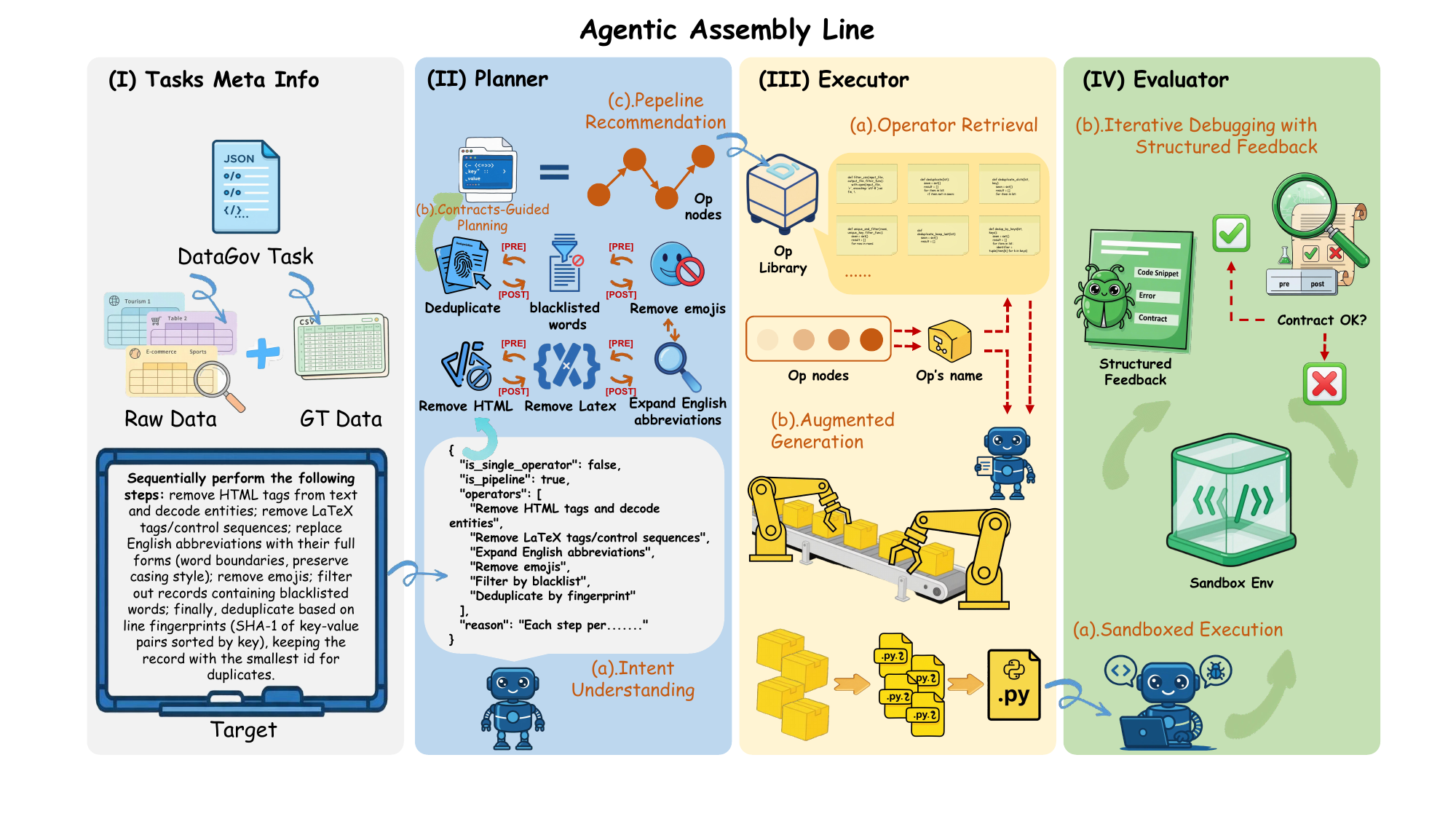}
    \caption{An overview of the Agentic Assembly Line. 
            The process progresses from intent understanding to contract-guided planning.
            Unlike traditional layouts, this figure spans roughly 1.5 columns, allowing the caption to sit snugly on the right side, creating a wrapping effect efficiently.}
    \label{fig:architecture}
\end{figure*}

To address the challenges of automating data governance, we introduce DataGovAgent, a novel multi-agent framework designed to interpret natural language instructions and autonomously orchestrate a DAG of data governance operations~\citep{guo2024multi,tran2025multi}.
The entire process, which we term \textbf{NL2GovDAG} (from natural language to a data-governance DAG), is operationalized through what we call an \textbf{Agentic Assembly Line}—a deterministic multi-agent workflow where specialized agents collaborate sequentially (Planner → Executor → Evaluator). Each step is governed by formal \textbf{governance contracts}, which are (pre, post) specifications that define input requirements and output guarantees for each operation. When execution fails, the system employs \textbf{feedback-driven debugging}, an iterative refinement process where agents reflect on their execution failures and generate targeted fixes based on contract violations and error analysis.

\subsection{Architectural Overview}
DataGovAgent employs an Agentic Assembly Line architecture (see Figure \ref{fig:architecture}), enabling systematic decomposition and execution of data governance tasks through multi-agent collaboration.

\subsection{Specialized Agent Roles}
Our framework is instantiated by three core agent roles—the Planner, Executor, and Evaluator~\citep{xu2024llm4workflow,chen2025position}. Their functions are orchestrated within a deterministic task chain, ensuring a structured progression from high-level intent to a verified, executable output.

Anchored in the data schema and representative samples, the \textbf{\textit{Planner}} uses few-shot prompting to align user intent with the actual data and to assess feasibility; it then extracts verifiable \textbf{governance contracts} that formalize each operator as a (pre, post) tuple~\citep{liu2024propertygpt,godboley2025sol}. Under these contracts, the Planner synthesizes an \textbf{initial DAG of abstract operators} such that the post-condition of each step satisfies the pre-condition of the next; when a constraint is not met, it inserts minimal repairs (for example, type casting or missing-value imputation) to ensure the pipeline is topologically coherent and executable. 

For each DAG node, the \textbf{\textit{Executor}} employs retrieval-augmented generation \citep{parvez2021retrieval,trirat2025automlagent}: it first retrieves the most relevant, validated operators from a curated library \citep{dataflow2025} and then injects their descriptions and snippets as dynamic in-context exemplars to guide code synthesis, yielding Python implementations that are tailored to the task while adhering to established best practices, thereby reducing hallucinations and improving reuse. 

The \textbf{\textit{Evaluator}} executes the generated code in a restricted sandbox; upon any failure or noncompliance, it captures the offending code region, full error messages, and stack traces, and ties them to the violated contracts to produce targeted revision advice. This feedback drives a guided correction loop until each operator is both runnable and contract-compliant, providing progressive validation on both construction and execution paths of the GovDAG. Implementation details and prompt templates are provided in \textbf{Appendix} \ref{appendix:agent roles}.

% ----------------------------------------------------------------------
\section{Experimental Setup}
\label{sec:experiments}

\paragraph{Benchmark and Metrics}
We evaluate on \textbf{DataGovbench-150}, which comprises 150 real-world data governance tasks. Each task includes a natural language description, raw datasets, and an executable script for automated scoring. The tasks are categorized into \textbf{Operator-level} (single-step operations) and \textbf{DAG-level} (complex multi-step workflows). We adopt the metrics detailed in Table \ref{tab:evaluation_metrics_final}, primarily focusing on Task Success Rate (TSR) for quality and Average Debug Iterations (ADI) for efficiency.

\paragraph{Baselines}
We compare DataGovAgent against two categories of baselines:
(1) \textbf{Single-Model Baselines}: Direct solution generation using state-of-the-art LLMs (e.g., GPT-5, GPT-4o) without agentic collaboration.
(2) \textbf{Agent Frameworks}: Representative multi-agent frameworks, specifically ChatDev \citep{qian-etal-2024-chatdev} and CAMEL \citep{li2023camel}, adapted for data governance tasks.

\section{Benchmark Results}
\label{sec:BENCHMARKRESULTS}

\subsection{Performance of Single-Model Baselines}

We evaluated the performance of single-model baselines on the operator-level tasks. The results are presented in Table \ref{tab:opensource_performance} and Table \ref{tab:closedsource_performance}.

\begin{table}[t!]
\centering
\caption{Performance of \textbf{Open-Source} Models on DataGovbench (Operator-Level)}
\label{tab:opensource_performance}
\renewcommand{\arraystretch}{1.2}
% 1. 减小列间距，这样可以让原本的字体在缩放后显得大一点
\setlength{\tabcolsep}{2pt} 
% 2. 使用 resizebox 强制将表格宽度限制在当前栏宽 (\linewidth) 内
\resizebox{\linewidth}{!}{
\begin{tabular}{
  l
  S[table-format=2.2]  % ATS
  S[table-format=2.2]  % TSR
  S[table-format=2.2]  % CRR
  S[table-format=2.2]  % Avg.
  S[table-format=6.2]  % Tokens
  S[table-format=4.2]  % Gen Time
  S[table-format=4.2]  % Exec Time
}
\toprule
\textbf{Model} &
{\textbf{ATS$\uparrow$}} &
{\textbf{TSR$\uparrow$}} &
{\textbf{CRR$\uparrow$}} &
{\textbf{Avg.\ Score$\uparrow$}} &
{\makecell{\textbf{Avg.} \textbf{Tokens}$\downarrow$}} &
{\makecell{\textbf{Generation}\\\textbf{Time (s)$\downarrow$}}} &
{\makecell{\textbf{Execution}\\\textbf{Time (s)$\downarrow$}}}\\
\midrule
Qwen3-235b-a22b & 34.73 & 46.00 & 69.00 & 49.91 & 950.68  & 1335.47 & 519.87 \\
Qwen2.5-coder   & 27.99 & 38.00 & 58.00 & 41.33 & \cellcolor{lowLite}589.57 & 1039.39 & 81.26 \\
Qwen3-coder     & \cellcolor{highDeep}38.74 & \cellcolor{highDeep}48.00 & 67.00 & \cellcolor{highLite}51.25 & 732.50 & \cellcolor{lowLite}185.07 & 122.60 \\
DeepSeek-V3     & \cellcolor{highLite}35.68 & \cellcolor{highLite}47.00 & \cellcolor{highLite}74.00 & \cellcolor{highDeep}52.23 & 680.51 & 1663.45 & 572.13 \\
Llama-3-70B     & 26.87 & 35.00 & 49.00 & 36.96 & \cellcolor{lowDeep}536.03 & \cellcolor{lowDeep}140.12 & \cellcolor{lowLite}72.48 \\
Llama-4-scout   & 14.88 & 23.00 & 37.00 & 24.96 & 702.50  & 618.06  & 151.65 \\
Mistral-7B      & 10.41 & 15.00 & 27.00 & 17.47 & 715.78 & 525.99 & 87.74 \\
% \midrule
Gemma-3-27B     & 29.62 & 43.00 & \cellcolor{highDeep}76.00 & 49.54 & 1425.84 & 4042.13 & \cellcolor{lowDeep}60.92 \\
Phi4     & 23.24 & 32.00 & 42.00 & 32.41 & 982.37 & 1642.61 & 98.73 \\
\bottomrule
\end{tabular}
} % resizebox 结束的大括号
\end{table}

% ===============================
\begin{table}[t!]
\centering
\caption{Performance of \textbf{Closed-Source} Models on DataGovbench (Operator-Level)}
\label{tab:closedsource_performance}
\renewcommand{\arraystretch}{1.2}
% 1. 减小列间距，为文字争取更多空间
\setlength{\tabcolsep}{2pt}
% 2. 使用 resizebox 自动缩放至栏宽
\resizebox{\linewidth}{!}{
\begin{tabular}{
  l
  S[table-format=2.2]
  S[table-format=2.2]
  S[table-format=2.2]
  S[table-format=4.2]
  S[table-format=4.2]
  S[table-format=4.2]
  S[table-format=3.2]
}
\toprule
\textbf{Model} &
{\textbf{ATS$\uparrow$}} &
{\textbf{TSR$\uparrow$}} &
{\textbf{CRR$\uparrow$}} &
{\textbf{Avg.\ Score$\uparrow$}} &
{\makecell{\textbf{Avg.} \textbf{Tokens}$\downarrow$}} &
{\makecell{\textbf{Generation}\\\textbf{Time (s)$\downarrow$}}} &
{\makecell{\textbf{Execution}\\\textbf{Time (s)$\downarrow$}}}\\
\midrule
GPT-5                & \cellcolor{highLite}40.98 & \cellcolor{highDeep}49.00 & \cellcolor{highLite}81.00 & \cellcolor{highDeep}56.99 & 3706.21 & 3069.44 & 598.73 \\
GPT-4o               & 32.04 & 41.00 & 56.00 & 43.01 & \cellcolor{lowDeep}555.26 & \cellcolor{lowDeep}431.85 & \cellcolor{lowDeep}29.72 \\
o4-mini              & \cellcolor{highDeep}41.47 & \cellcolor{highDeep}49.00 & 68.00 & 52.82 & 1510.68 & 1127.16 & 167.28 \\
o1                   & 32.50 & 41.00 & 74.00 & 49.17 & 1908.54 & 3916.70 & 35.55 \\
o3                   & 34.48 & 45.00 & 63.00 & 47.49 & 1415.08 & 1291.82 & \cellcolor{lowLite}35.16 \\
% \midrule
Claude-4-sonnet      & 36.75 & 46.00 & \cellcolor{highDeep}85.00 & 55.92 & 1672.91 & 3149.83 & 229.70 \\
Claude-4-opus        & 38.30 & 47.00 & 79.00 & 54.77 & 1390.04 & 3298.22 & 158.85 \\
Gemini-2.5-flash     & 40.26 & \cellcolor{highLite}48.00 & 80.00 & \cellcolor{highLite}56.09 & 5234.30 & 5727.56 & 355.65 \\
Grok-3               & 35.41 & 44.00 & 71.00 & 50.14 & \cellcolor{lowLite}688.51 & \cellcolor{lowLite}811.22 & 685.25 \\
Grok-4               & 36.90 & 44.00 & 67.00 & 49.30 & 4575.07 & 7700.30 & 406.62 \\
Kimi-K2-instruct    & 39.52 & \cellcolor{highDeep}49.00 & 70.00 & 52.84 & 721.16 & 864.21 & 652.62 \\
\bottomrule
\end{tabular}
}
\end{table}

From the performance of the single-model baselines, we observe the following:

\textbf{Significant Performance Ceiling:} Even the most powerful closed-source models, such as GPT-5 and Claude4-sonnet, fail to exceed a 50\% TSR in a single-round code generation setting. This indicates that the tasks in DataGovbench are considerably challenging and difficult to solve perfectly with a single code generation attempt.

\textbf{Runnable Does Not Equal Correct:} Many models, such as \textbf{Claude4-sonnet}, exhibit a very high Code Runnability Rate (CRR $>$ 80\%), yet their TSR is significantly lower. This reveals a critical issue: models can generate syntactically correct code, but the logic of this code does not necessarily meet the business objectives of the task.

\textbf{Potential of Open-Source Models:} Leading open-source code models, represented by \textbf{DeepSeek-V3}, can match or even surpass some closed-source models in TSR. This demonstrates their strong potential in the data science domain.

Building upon this, we have also systematically evaluated these models on the more challenging DAG-Level tasks. Unlike single-operator tasks, DAG tasks require the model to generate a \textbf{complete data processing workflow }in a single pass. This involves: 1) correctly decomposing the task into sub-tasks, 2) organizing them in a logical execution order, 3) ensuring correct dependency passing between steps, and 4) producing a final output that meets the specified business objectives. Due to the significant increase in complexity, the Avg. Score on DAG-Level tasks is generally lower than that on Operator-Level tasks. 

\begin{table}[t!]
\centering
\caption{Performance of \textbf{Open-Source} Models on DataGovbench (DAG-Level)}
\label{tab:opensource_performance_dag}
\renewcommand{\arraystretch}{1.2}
\setlength{\tabcolsep}{2pt}
\resizebox{\linewidth}{!}{
\begin{tabular}{
  l
  S[table-format=2.2]  % ATS
  S[table-format=2.2]  % TSR
  S[table-format=2.2]  % CRR
  S[table-format=2.2]  % Avg.
  S[table-format=4.2]  % Tokens
  S[table-format=6.2]  % Gen Time
  S[table-format=3.2]  % Exec Time
}
\toprule
\textbf{Model} &
{\textbf{ATS$\uparrow$}} &
{\textbf{TSR$\uparrow$}} &
{\textbf{CRR$\uparrow$}} &
{\textbf{Avg.\ Score$\uparrow$}} &
{\makecell{\textbf{Avg.} \textbf{Tokens}$\downarrow$}} &
{\makecell{\textbf{Generation}\\\textbf{Time (s)$\downarrow$}}} &
{\makecell{\textbf{Execution}\\\textbf{Time (s)$\downarrow$}}}\\
\midrule
Qwen3-235b-a22b      & \cellcolor{highLite}25.64 & \cellcolor{highLite}38.00 & \cellcolor{highLite}50.00 & \cellcolor{highLite}37.88 & 3005.22 & 7339.20 & 81.43 \\
Qwen2.5-coder        & 12.11 & 26.00 & 30.00 & 22.70 & \cellcolor{lowLite}738.68 & 852.36 & 28.23 \\
Qwen3-coder          & 20.87 & 36.00 & 48.00 & 34.96 & 1075.36 & \cellcolor{lowDeep}77.32 & 370.27 \\
DeepSeek-V3          & \cellcolor{highDeep}28.65 & \cellcolor{highDeep}56.00 & \cellcolor{highDeep}72.00 & \cellcolor{highDeep}52.22 & 983.70 & 1098.90 & 305.99 \\
Llama-3-70B          &  8.07 & 10.00 & 16.00 & 11.36 & \cellcolor{lowDeep}723.08 & 284.43 & 221.09 \\
Llama-4-scout        &  7.35 & 12.00 & 22.00 & 13.78 & 864.16 & 435.08 & \cellcolor{lowDeep}10.39 \\
Mistral-7B           &  7.10 & 18.00 & 20.00 & 15.03 & 897.88 & \cellcolor{lowLite}261.90 & 230.13 \\
% \midrule
Gemma-3-27B          & 11.31 & 20.00 & 38.00 & 23.10 & 1671.34 & 2412.24 & 19.06 \\
Phi-4                &  6.73 & 20.00 & 28.00 & 18.24 & 1081.94 & 929.29 & \cellcolor{lowLite}18.35 \\
\bottomrule
\end{tabular}
}
\vspace{-10pt}
\end{table}

\begin{table}[t!]
\centering
\caption{Performance of \textbf{Closed-Source} Models on DataGovbench (DAG-Level)}
\label{tab:closedsource_performance_dag}
\renewcommand{\arraystretch}{1.2}
\setlength{\tabcolsep}{2pt}
\resizebox{\linewidth}{!}{
\begin{tabular}{
  l
  S[table-format=2.2]  % ATS
  S[table-format=2.2]  % TSR
  S[table-format=2.2]  % CRR
  S[table-format=2.2]  % Avg.
  S[table-format=4.2]  % Tokens
  S[table-format=6.2]  % Gen Time
  S[table-format=3.2]  % Exec Time
}
\toprule
\textbf{Model} &
{\textbf{ATS$\uparrow$}} &
{\textbf{TSR$\uparrow$}} &
{\textbf{CRR$\uparrow$}} &
{\textbf{Avg.\ Score$\uparrow$}} &
{\makecell{\textbf{Avg.} \textbf{Tokens}$\downarrow$}} &
{\makecell{\textbf{Generation}\\\textbf{Time (s)$\downarrow$}}} &
{\makecell{\textbf{Execution}\\\textbf{Time (s)$\downarrow$}}}\\
\midrule
GPT-5                & 27.18 & 46.00 & \cellcolor{highDeep}86.00 & 53.06 & 6086.82 & 7121.52 & 310.05 \\
GPT-4o               & 18.68 & 38.00 & 50.00 & 35.56 & \cellcolor{lowDeep}754.82 & \cellcolor{lowDeep}276.54 & \cellcolor{lowLite}52.94 \\
o4-mini              & \cellcolor{highLite}31.86 & \cellcolor{highDeep}56.00 & 74.00 & \cellcolor{highLite}53.95 & 2075.26 & 971.14 & 91.31 \\
o1                   & 27.79 & 52.00 & \cellcolor{highLite}80.00 & 53.26 & 2574.00 & 3270.06 & \cellcolor{lowDeep}15.68 \\
o3                   & 31.22 & 46.00 & 64.00 & 47.07 & 2027.76 & 1410.07 & 85.07 \\
% \midrule
Claude-4-sonnet      & \cellcolor{highDeep}34.77 & \cellcolor{highLite}54.00 & 76.00 & \cellcolor{highDeep}54.92 & 1890.82 & 2007.23 & 143.01 \\
Claude-4-opus        & 20.41 & 34.00 & 50.00 & 34.80 & 1759.84 & 2443.04 & 74.24 \\
Gemini-2.5-flash     & 25.40 & 44.00 & 68.00 & 45.80 & 7383.40 & 2457.91 & 295.21 \\
Grok-3               & 27.45 & 46.00 & 62.00 & 45.15 & \cellcolor{lowLite}854.72 & \cellcolor{lowLite}626.97 & 194.63 \\
Grok-4               & 31.38 & 50.00 & 66.00 & 49.13 & 5537.42 & 4706.45 & 277.36 \\
Kimi-K2-instruct    & 20.60 & 30.00 & 34.00 & 28.20 & 1107.94 & 758.61 & 80.78 \\
\bottomrule
\end{tabular}
}
\end{table}

Tables \ref{tab:opensource_performance_dag} and \ref{tab:closedsource_performance_dag} summarize the baseline results for the open-source and closed-source models.

\textbf{Top-Tier Open-Source Models Rival Closed-Source Counterparts:} On DAG tasks, the leading open-source model, DeepSeek-V3 \citep{deepseekv3}, achieved a 56.00 Task TSR. This performance not only leads the open-source field but also matches the top-performing closed-source model, o4-mini (56.00 TSR), while outperforming other powerful models like GPT-5 (46.00). This strongly indicates that leading open-source code models are highly competitive for handling complex, end-to-end data science workflows.

\begin{figure}[!h]
    \centering
    % 上图：performance
    \includegraphics[width=1\linewidth]{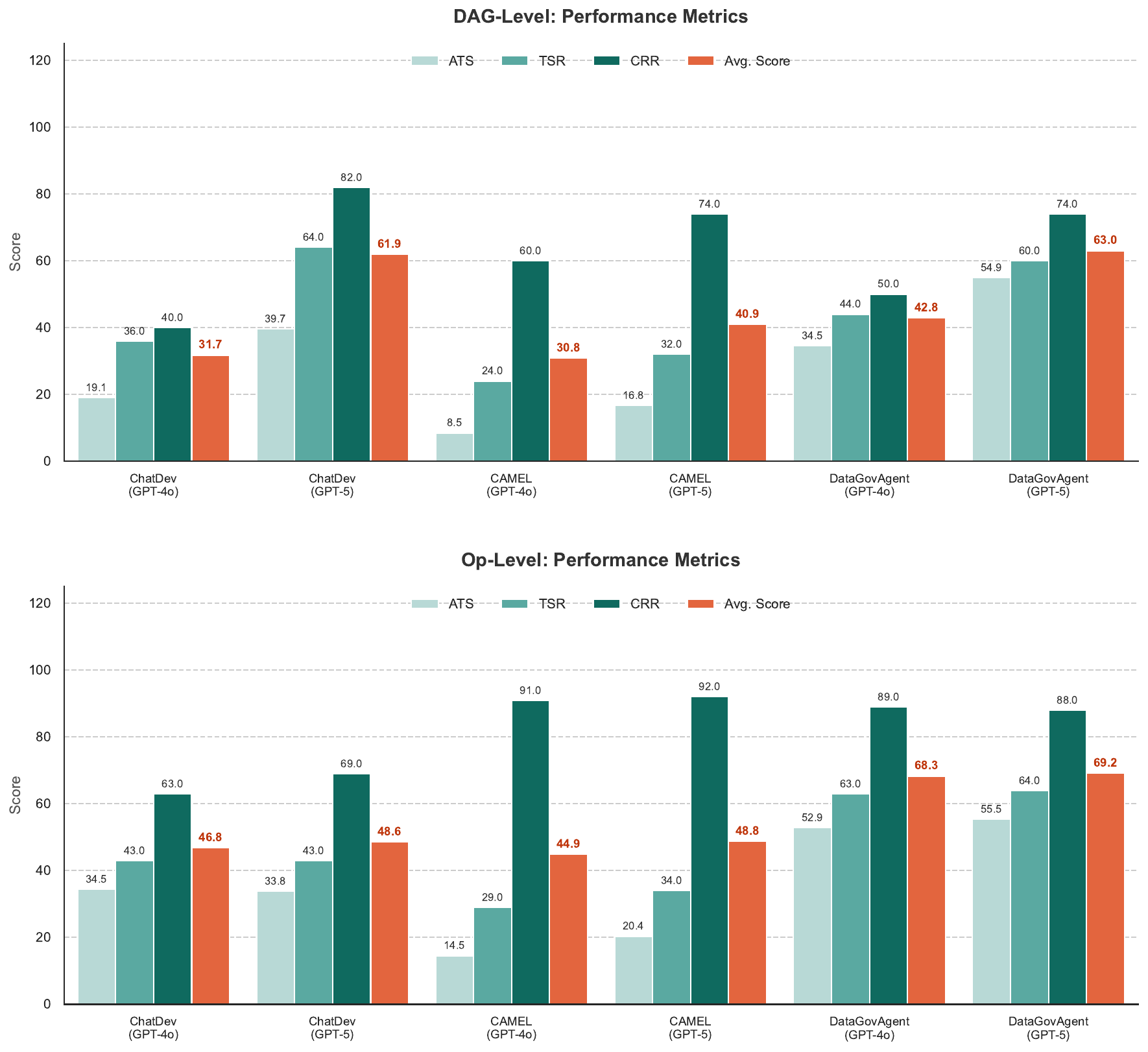}
    
    \vspace{0.5em}

    % 下图：efficiency
    \includegraphics[width=1\linewidth]{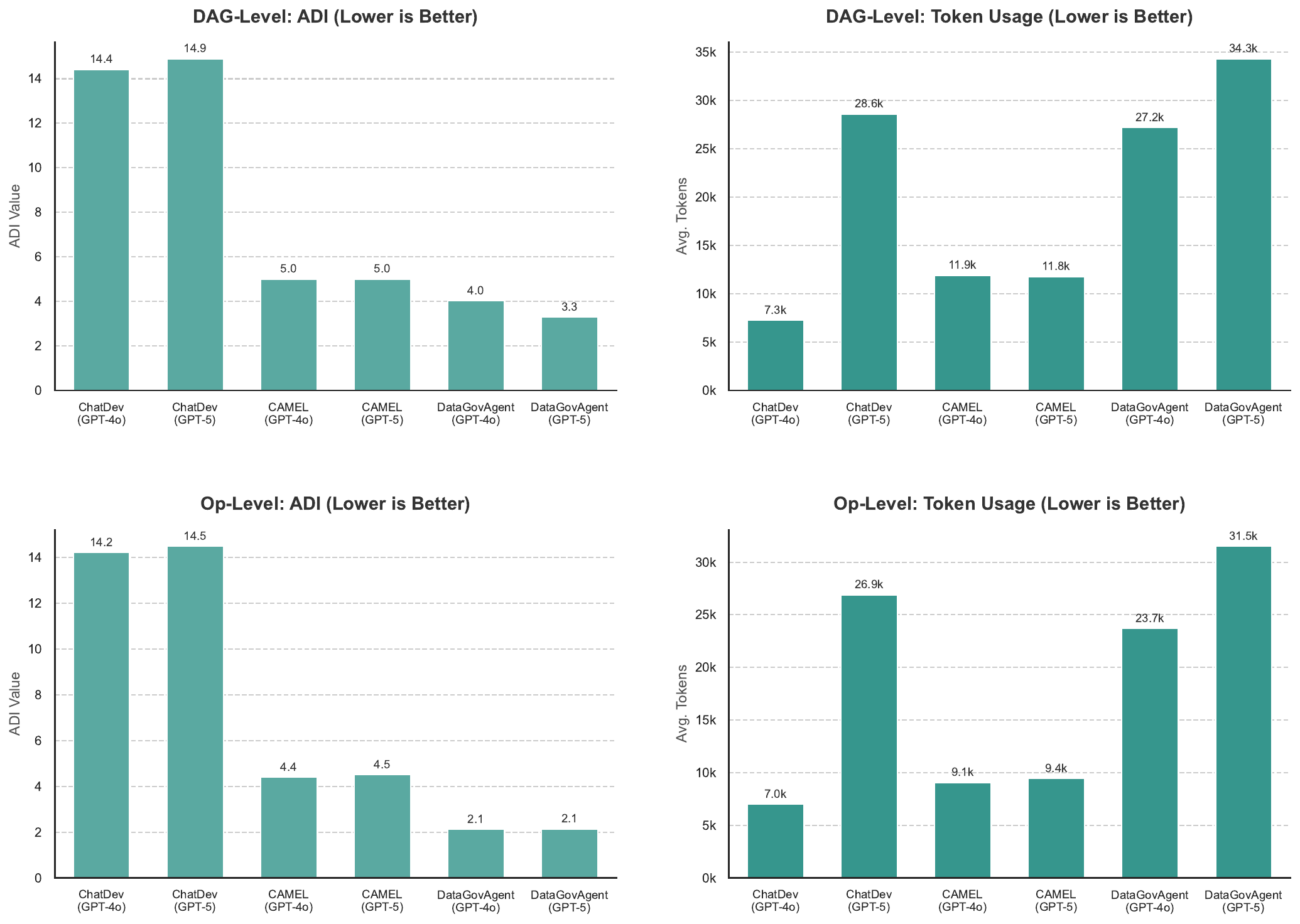}
    
    \caption{Performance of Agent Framework Baselines}
    \label{tab:govbench_split}
\end{figure}

\textbf{Performance Divergence Among Closed-Source Models:} Within the closed-source camp, models exhibit different strengths. o4-mini demonstrates superior task-solving ability with the highest TSR. In contrast, Claude4-sonnet excels in ATS and Average Score, suggesting its generated code has higher overall quality and completeness. This reflects different optimization priorities among proprietary models.

The \textbf{"Runnable $\ne$ Correct"} Gap Is More Pronounced: In complex DAG tasks, the disparity between a high CRR and a low TSR is even more significant (e.g., GPT‑5). For instance, GPT-5 shows an 86 CRR but only a 46 TSR. This reaffirms that generating syntactically correct complex workflows does not guarantee logical adherence to business objectives. Notably, the top-performing DeepSeek-V3 has a smaller gap between its CRR (72) and TSR (56), potentially indicating a better alignment between its code's runnability and its logical correctness.

\textbf{A Clear Trade-off Between Efficiency and Performance Persists:} The GPT-4o model demonstrates high generation efficiency, with the lowest token count and generation time among closed-source models. However, its 38.00 TSR is considerably lower than that of top-tier models. This highlights a clear trade-off between speed and accuracy when handling complex tasks, where some models achieve higher accuracy at a greater computational cost, while others are optimized for a balance between efficiency and performance.

%  =======================================================================================================================================
\subsection{Performance of Agent Framework Baselines}
We evaluated the ChatDev and CAMEL frameworks on DataGovbench by pairing them with powerful GPT-4o and GPT-5 models in Figure \ref{tab:govbench_split}.

\textbf{Closing the Runnable–Correct Gap with Contracts and Feedback:} On DataGovbench, DataGov‑Agent consistently turns runnability into business‑correct solutions more efficiently than generic agent frameworks. On DAG‑level tasks, although ChatDev+GPT‑5 attains the top TSR (64), DataGov‑Agent+GPT‑5 delivers higher average quality (ATS 54.91 vs. 39.67; Avg. Score 62.97 vs. 61.89), requires 4.5× fewer debug iterations (ADI 3.29 vs. 14.89). On operator‑level tasks, DataGov‑Agent+GPT‑5 leads in TSR/ATS/Avg. Score (64/55.47/69.15) and shows the strongest alignment between runnability and correctness (A=TSR/CRR=0.73 vs. 0.62 for ChatDev and 0.37 for CAMEL), indicating that contracts and meta‑cognitive feedback effectively convert CRR into TSR. More detailed analysis in \textbf{Appendix} 
\ref{appendix:details}.

\section{Conclusion}
\label{sec:conclusion}
We present DataGovbench, the first benchmark designed to comprehensively stress-test large language model agents on real-world data governance tasks. DataGovbench offers two main contributions: it provides a two-tiered task suite that spans from atomic operators to multi-step DAG pipelines, and for each task, it incorporates unique evaluation logic and scoring metrics. Furthermore, our proposed DataGovAgent achieves SOTA performance on this new benchmark, significantly outperforming existing agent frameworks on complex governance pipelines.

\newpage
\section{Limitations}
\label{sec:limitations}
\paragraph{Scalability of Benchmark Construction.}
Constructing our benchmark involved a rigorous human-in-the-loop process to ensure the reliability of task objectives and ground-truth DAGs. While this manual curation is essential for a fair evaluation, it inherently limits the dataset's scalability. Expanding the benchmark to the scale required for extensive training remains a labor-intensive challenge, which we aim to address in future work through semi-automated problem synthesis.

\paragraph{Efficiency and Semantic Alignment.}
The iterative nature of our \textit{Agentic Assembly Line}, while crucial for robustness, inevitably incurs higher token consumption and latency compared to single-pass models. This computational overhead represents a trade-off between reliability and resource efficiency. Furthermore, while our governance contracts effectively enforce structural correctness, they may not fully capture the subtle nuances of highly ambiguous user instructions. In rare cases, the agent might generate a compliant but semantically misaligned DAG, suggesting a need for future mechanisms that incorporate proactive human clarification.

\label{sec:limitation}

\bibliography{custom}

\onecolumn
\newpage
\twocolumn

\appendix
\section{Appendix}
\label{appendix:a}

\subsection{Benchmark Comparison Table}
\label{appendix:related works table}

\begin{table*}[ht!]
  \centering
  \caption{Evolution of Code\,\&\,Agent Benchmarks
           (textual overview; corresponding visual examples
           are shown in Figure~\ref{fig:benchdemo_grid}).}
  \label{tab:bench_comparison_text}
  \small                 % 字体稍微缩小，行距更紧凑
  \setlength{\extrarowheight}{2pt}
  \begin{tabularx}{\textwidth}{
    >{\raggedright\arraybackslash}p{2.8cm}   % Benchmark
    >{\raggedright\arraybackslash}p{2.8cm}   % Evaluation Scope
    X                                        % Key Features (自动换行)
    X                                        % Methodological Focus
  }
    \toprule
    \textbf{Benchmark} &
    \textbf{Evaluation Scope} &
    \textbf{Key Features} &
    \textbf{Methodological Focus} \\
    \midrule
    DS-1000            & Snippet-level   & Code generation for data-science libraries (NumPy, Pandas) & Basic code completion \\
    DA-Code            & Task-level      & Extends DS-1000 with an \emph{interactive} execution environment & Interactive problem solving \\
    DataSciBench       & Workflow-level  & Systematic LLM-agent evaluation with 25 multidimensional metrics & Complete data-science pipelines \\
    ScienceAgentBench  & Domain-specific & Rigorous assessment for data-driven scientific discovery & Scientific research workflows \\
    HumanEval Pro      & Reasoning-focused & Self-invoking code generation with progressive reasoning & Advanced reasoning capabilities \\
    LiveBench          & Methodology-focused & Dynamic benchmark that mitigates dataset contamination & Evaluation robustness \\
    \rowcolor{gray!15}%
    DataGovbench           & Hierarchical (Operator \& DAG-level) & 150 realistic tasks; reversed-objective noise; multi-metric scoring (ATS/TSR/CRR) & End-to-end data-governance pipeline evaluation \\
    \bottomrule
  \end{tabularx}
\end{table*}

\begin{figure}[htbp]
  \centering
  
  % ---------- 第 1 行 ----------
  \begin{subfigure}[t]{0.45\linewidth}
    \centering
    \includegraphics[width=\linewidth]{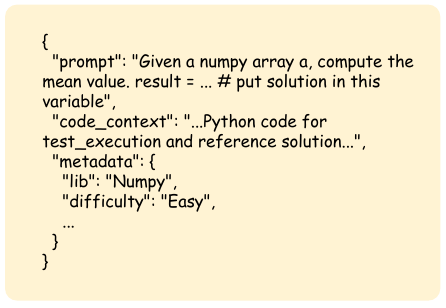}
    \caption{DS-1000}
  \end{subfigure}\hfill
  \begin{subfigure}[t]{0.45\linewidth}
    \centering
    \includegraphics[width=\linewidth]{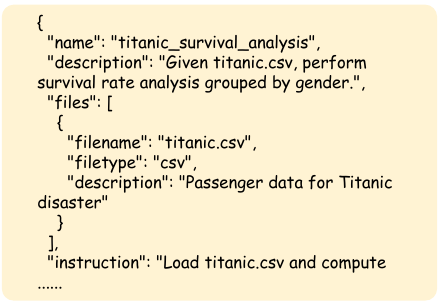}
    \caption{DA-Code}
  \end{subfigure}\\[4pt]
  
  % ---------- 第 2 行 ----------
  \begin{subfigure}[t]{0.45\linewidth}
    \centering
    \includegraphics[width=\linewidth]{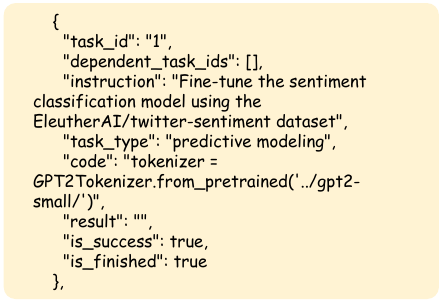}
    \caption{DataSciBench}
  \end{subfigure}\hfill
  \begin{subfigure}[t]{0.45\linewidth}
    \centering
    \includegraphics[width=\linewidth]{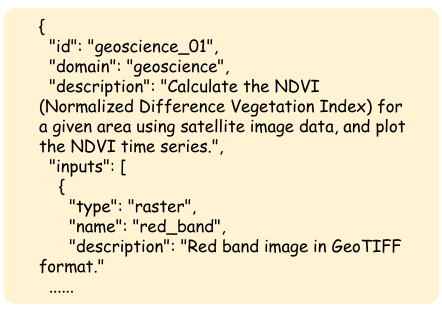}
    \caption{ScienceAgentBench}
  \end{subfigure}\\[4pt]
  
  % ---------- 第 3 行 ----------
  \begin{subfigure}[t]{0.45\linewidth}
    \centering
    \includegraphics[width=\linewidth]{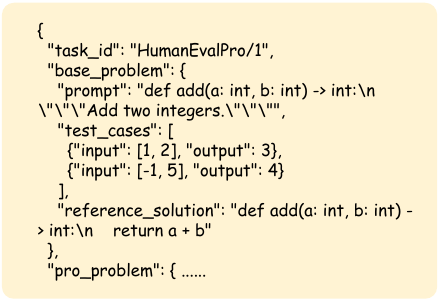}
    \caption{HumanEval Pro}
  \end{subfigure}\hfill
  \begin{subfigure}[t]{0.45\linewidth}
    \centering
    \includegraphics[width=\linewidth]{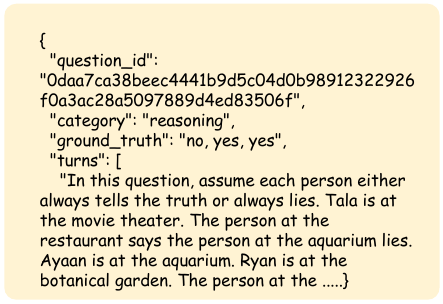}
    \caption{LiveBench}
  \end{subfigure}\\[4pt]
  
  % ---------- 第 4 行 ----------
  \begin{subfigure}[t]{0.45\linewidth}
    \centering
    \includegraphics[width=\linewidth]{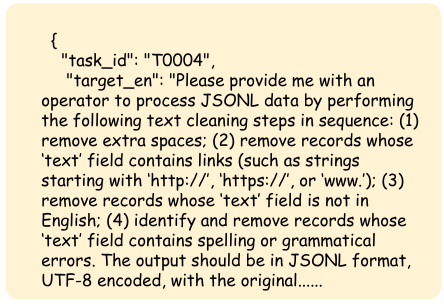}
    \caption{DataGovbench}
  \end{subfigure}
  
  \caption{BenchDemo visual examples in a compact layout.}
  \label{fig:benchdemo_grid}
\end{figure}

\subsection{Tasks Eval}
\label{appendix:Tasks Eval}

\begin{table}[!t]
\centering
\caption{Evaluation Metrics (Compact View)}
\label{tab:op_eval_metrics}
% 使用 tabularx 自动计算宽度
\begin{tabularx}{\linewidth}{@{} >{\raggedright\arraybackslash}p{0.3\linewidth} X @{}}
\toprule
\textbf{Task \& Metric} & \textbf{Description} \\
\midrule

\textbf{Filtering} \newline
\textit{Metric: F1 Score}
& Measures the balance of precision and recall in correctly identifying and removing erroneous or unwanted data rows. \\
\addlinespace[6pt] % 增加行间距，区分不同条目

\textbf{Refinement} \newline
\textit{Metric: Accuracy}
& Assesses the correctness of data transformations, such as standardizing date formats or parsing text. \\
\addlinespace[6pt]

\textbf{Imputation} \newline
\textit{Metric: Completion Rate}
& Evaluates the model's effectiveness in correctly filling in missing or null values based on the ground truth. \\
\addlinespace[6pt]

\textbf{Deduplication} \newline
\textit{Metric: Consistency}
& Measures the success in identifying and removing duplicate records or ensuring data consistency. \\
\addlinespace[6pt]

\textbf{Data Integration} \newline
\textit{Metric: Integration Acc.}
& Assesses how well data from different sources is merged, handling schema mismatches and conflicts. \\
\addlinespace[6pt]

\textbf{Classification} \newline
\textit{Metric: Accuracy/F1}
& Uses standard classification metrics to evaluate the correctness of labels assigned to data records. \\

\bottomrule
\end{tabularx}
\end{table}
See evaluation Metrics for Operator-Level Task Categories in DataGovbench in Table \ref{tab:op_eval_metrics}.

\subsection{Benchmark Examples}
This part shows details of sample tasks across six Operator-level tasks and DAG tasks, including natural language task objectives, evaluation script snippets and dataset samples.

\begin{table*}
\begin{tcolorbox}[colframe=myblue1, colback=mygray1, coltitle=white, title=Filtering Task Objective, arc=1mm, boxrule=0.8mm]
Please write an operator to process jsonl files, filtering out text entries that contain blocked words (such as offensive, vulgar, or obscene words) in the text field. Each record is a JSON object, and it is necessary to check whether its text field contains blocked words. After filtering out these records, output a new JSONL file, keeping the field structure unchanged and encoded in UTF-8.
\end{tcolorbox}
\end{table*}
    
\begin{table*}
\begin{tcolorbox}[colframe=myblue1, colback=mygray1, coltitle=white, title=Filtering Task Eval Code, 
                 arc=1mm, boxrule=0.8mm, breakable]
\begin{lstlisting}[language=Python]
def evaluate(expected_path, processed_path):
    expected = load_jsonl(expected_path)
    processed = load_jsonl(processed_path)

    expected_ids = set(entry['id'] for entry in expected)
    processed_ids = set(entry['id'] for entry in processed)

    true_positives = len(expected_ids & processed_ids)
    predicted_total = len(processed_ids)
    gold_total = len(expected_ids)

    precision = true_positives / predicted_total if predicted_total > 0 else 0.0
    recall = true_positives / gold_total if gold_total > 0 else 0.0
    f1 = (2 * precision * recall) / (precision + recall) if precision + recall > 0 else 0.0

    result = {"eval_score": f"{f1:.4f}"}
    print(result)
\end{lstlisting}
\end{tcolorbox}
\end{table*}

\begin{table*}
\begin{tcolorbox}[
    colframe=myblue1, 
    colback=mygray1, 
    coltitle=white, 
    title=Filtering Task Dataset Samples, 
    % sharp corners=southwest, 
    arc=2mm, 
    boxrule=0.8mm,
    width=\textwidth
]
\centering
\begin{tabular}{|p{0.45\textwidth}|p{0.45\textwidth}|}
\hline
\textbf{Raw Data} & \textbf{GT Data} \\
\hline
% Left Column Data
\{ "id": 0, "text": "The customer service team is committed to providing exceptional assistance to all clients."\} \newline
\{ "id": 1, "text": "Customer service is dedicated to addressing and resolving client inquiries efficiently."\} \newline
\{ "id": 2, "text": "The refund policy stipulates that all requests must be submitted within thirty days of purchase."\} \newline
\{ "id": 3, "text": "The refund policy allows customers to return products within 30 days for a full refund."\} \newline
\{ "id": 4, "text": "The product warranty is valid for a period of one year from the date of purchase."\} \newline 
{"id": 32, "text": "Fix this crap service now."} \newline 
{"id": 43, "text": "Fix this damn service now."} \newline 
{"id": 58, "text": "This whole thing is total bitch."} \newline 
{"id": 33, "text": "Why the bitch is my order delayed?"} \newline 
{"id": 83, "text": "Why the bastard is my order delayed?"} \newline
& 
% Right Column Data
\{ "id": 0, "text": "The customer service team is committed to providing exceptional assistance to all clients."\} \newline
\{ "id": 1, "text": "Customer service is dedicated to addressing and resolving client inquiries efficiently."\} \newline
\{ "id": 2, "text": "The refund policy stipulates that all requests must be submitted within thirty days of purchase."\} \newline
\{ "id": 3, "text": "The refund policy allows customers to return products within 30 days for a full refund."\} \newline
\{ "id": 4, "text": "The product warranty is valid for a period of one year from the date of purchase."\} \newline \\
\hline
\end{tabular}
\end{tcolorbox}
\end{table*}

% \textbf{2. Refinement Task}

% Create a tcolorbox with title and content
\begin{table*}
\begin{tcolorbox}[colframe=myblue1, colback=mygray1, coltitle=white, title=Refinement Task Objective, arc=1mm, boxrule=0.8mm]
Please write an operator to process JSONL files and remove HTML tags from the text field. Each record is a JSON object, requiring detection of its text field and removal of all HTML tags (e.g., \textless p\textgreater, \textless a href=\textquotesingle url\textquotesingle \textgreater, etc.). Output a new JSONL file, retaining the field structure unchanged, encoded in UTF-8.
\end{tcolorbox}
\end{table*}

\begin{table*}
\begin{tcolorbox}[colframe=myblue1, colback=mygray1, coltitle=white, title=Refinement Task Eval Code, 
                 arc=1mm, boxrule=0.8mm, breakable]
\begin{lstlisting}[language=Python]
def evaluate(expected_path, processed_path, show_diff=5):
    expected = load_jsonl(expected_path)
    processed = load_jsonl(processed_path)

    total = len(expected)
    matched = 0
    mismatches = []

    for id_, exp_text in expected.items():
        proc_text = processed.get(id_)
        if proc_text is None:
            mismatches.append((id_, "missing", exp_text, ""))
        else:
            if normalize(proc_text) == normalize(exp_text):
                matched += 1
            else:
                mismatches.append((id_, "mismatch", exp_text, proc_text))

    accuracy = matched / total if total > 0 else 0.0

    result = {"eval_score": f"{accuracy:.4f}"}
    print(result)
\end{lstlisting}
\end{tcolorbox}   
\end{table*}

\begin{table*}
\begin{tcolorbox}[
    colframe=myblue1, 
    colback=mygray1, 
    coltitle=white, 
    title=Refinement Task Dataset Samples, 
    % sharp corners=southwest, 
    arc=2mm, 
    boxrule=0.8mm,
    width=\textwidth
]
\centering
\begin{tabular}{|p{0.45\textwidth}|p{0.45\textwidth}|}
\hline
\textbf{Raw Data} & \textbf{GT Data} \\
\hline
% Left Column Data
\{ "id": "id\_0001", "topic": "climate change", "text": "Climate change poses significant challenges to the global environment and necessitates urgent collective action." \} \newline
\{ "id": "id\_0002", "topic": "climate change", "text": "Climate change poses a significant threat to the stability of ecosystems worldwide." \} \newline
\{ "id": "id\_0003", "topic": "climate change", "text": "Climate change poses a significant threat to global ecosystems and human societies." \} \newline
\{ "id": "id\_0004", "topic": "climate change", "text": "Climate change poses a significant threat to global ecosystems and human societies." \} \newline
\{ "id": "id\_0005", "topic": "climate change", "text": "Climate change presents a significant challenge that requires immediate global attention and action." \}
& 
% Right Column Data
\{ "id": "id\_0001", "topic": "climate change", "text": "Climate change poses significant challenges to the global environment and necessitates urgent collective action." \} \newline
\{ "id": "id\_0002", "topic": "climate change", "text": "Climate change poses a significant threat to the stability of ecosystems worldwide." \} \newline
\{ "id": "id\_0003", "topic": "climate change", "text": "Climate change poses a significant threat to global ecosystems and human societies." \} \newline
\{ "id": "id\_0004", "topic": "climate change", "text": "Climate change poses a significant threat to global ecosystems and human societies." \} \newline
\{ "id": "id\_0005", "topic": "climate change", "text": "Climate change presents a significant challenge that requires immediate global attention and action." \} \\
\hline
\end{tabular}
\end{tcolorbox}
\end{table*}

% \textbf{3. Imputation Task}

% Create a tcolorbox with title and content
\begin{table*}
\begin{tcolorbox}[colframe=myblue1, colback=mygray1, coltitle=white, title=Imputation Task Objective, arc=1mm, boxrule=0.8mm]
Need a data governance operator that uses the KNN algorithm (k=3) to impute missing values in a CSV file. 1. Input file: CSV (with header, comma-separated). 2. Supports numeric and one-hot encoded categorical variables. Encoding: UTF-8, no BOM.
\end{tcolorbox}

\begin{tcolorbox}[colframe=myblue1, colback=mygray1, coltitle=white, title=Imputation Task Eval Code, 
                 arc=1mm, boxrule=0.8mm, breakable]
\begin{lstlisting}[language=Python]
def evaluate(cand: pd.DataFrame,
             gt: pd.DataFrame,
             raw: pd.DataFrame) -> float:

    if cand.shape != gt.shape:
        fail(f"Mismatch in dimensions: Expected {gt.shape}, Actual {cand.shape}")
    if list(cand.columns) != list(gt.columns):
        fail("Column names or order do not match the reference")

    miss_mask = raw.isna()

    if cand[miss_mask].isna().any().any():
        fail("There are missing values that were not filled")

    diff = np.abs(cand[miss_mask].astype(float) - gt[miss_mask].astype(float))
    if (diff > ATOL).any().any():
        fail("The filled values do not match the reference (non-KNN imputation)")

    if not cand[~miss_mask].astype(float).equals(raw[~miss_mask].astype(float)):
        fail("The originally complete data has been modified")

    return 1.0
\end{lstlisting}
\end{tcolorbox}
\end{table*}

\begin{table*}
\begin{tcolorbox}[
    colframe=myblue1, 
    colback=mygray1, 
    coltitle=white, 
    title=Imputation Task Dataset Samples, 
    % sharp corners=southwest, 
    arc=2mm, 
    boxrule=0.8mm,
    width=\textwidth
]
\centering
\begin{tabular}{|p{0.45\textwidth}|p{0.45\textwidth}|}
\hline
\textbf{Raw Data} & \textbf{GT Data} \\
\hline
% Left Column Data
customer\_id, age, income, color\_blue, color\_green, color\_red & customer\_id, age, income, color\_blue, color\_green, color\_red \\
1, 22.0, 37110.61305675143, True, False, False & 1, 22.0, 37110.61305675143, 1.0, 0.0, 0.0 \\
2, 58.0, 55531.26176123748, False, False, True & 2, 58.0, 55531.26176123748, 0.0, 0.0, 1.0 \\
3, 52.0, 35616.760987565016, False, False, True & 3, 52.0, 35616.760987565016, 0.0, 0.0, 1.0 \\
4, 40.0, 63176.75451960909, True, , & 4, 40.0, 63176.75451960909, 1.0, 0.3333333333333333, 0.3333333333333333 \\
5, 40.0, 49251.11133520621, False, True, False & 5, 40.0, 49251.11133520621, 0.0, 1.0, 0.0 \\
6, 62.0, 47227.06454682109, False, , & 6, 62.0, 47227.06454682109, 0.0, 0.0, 0.3333333333333333 \\
7, 22.0, 39786.05683394088, True, False, False & 7, 22.0, 39786.05683394088, 1.0, 0.0, 0.0 \\
8, 54.0, 68338.12008011046, False, , True & 8, 54.0, 68338.12008011046, 0.0, 0.0, 1.0 \\
9, 28.0, 47682.05776896797, True, False, False & 9, 28.0, 47682.05776896797, 1.0, 0.0, 0.0 \\
10, 22.0, 43575.08266755339, False, False, True & 10, 22.0, 43575.08266755339, 0.0, 0.0, 1.0 \\
11, 45.0, , True, False, & 11, 45.0, 58632.88840075844, 1.0, 0.0, 0.0 \\
12, 68.0, 57984.63778330023, True, False, False & 12, 68.0, 57984.63778330023, 1.0, 0.0, 0.0 \\
13, , 55481.660965461175, True, False, & 13, 54.333333333333336, 55481.660965461175, 1.0, 0.0, 0.3333333333333333 \\
14, 57.0, 56190.98917393983, False, True, False & 14, 57.0, 56190.98917393983, 0.0, 1.0, 0.0 \\
15, 55.0, 56462.315045118245, , True, False & 15, 55.0, 56462.315045118245, 0.6666666666666666, 1.0, 0.0 \\
\hline
\end{tabular}
\end{tcolorbox}
\end{table*}

% \textbf{4. De-duplication Task}

% Create a tcolorbox with title and content
\begin{table*}
\begin{tcolorbox}[colframe=myblue1, colback=mygray1, coltitle=white, title=De-duplication Task Objective, arc=1mm, boxrule=0.8mm]
A data governance operator for incremental deduplication on \texttt{*.csv} / \texttt{*.jsonl}: 1. Historical baseline: \texttt{.jsonl} (already deduplicated, contains id, updated\_at, and business fields) 2. New incremental file: \texttt{.csv} (same structure) 3. Primary key: id 4. Deduplication rules: If the primary key exists in the baseline, ignore the incremental row; if not, append to the result set; For the same key but different business fields, keep the record with the latest updated\_at.
\end{tcolorbox}    
\end{table*}

\begin{table*}
\begin{tcolorbox}[colframe=myblue1, colback=mygray1, coltitle=white, title=De-duplication Task Eval Code, 
                 arc=1mm, boxrule=0.8mm, breakable]
\begin{lstlisting}[language=Python]
def compute_f1(
    gt_map: Dict[str, Dict],
    pred_rows: List[Dict],
) -> float:
    if not pred_rows:
        return 0.0

    tp_ids: Set[str] = set()
    fp = 0

    for row in pred_rows:
        rid = str(row.get("id", ""))
        if not rid:
            fp += 1
            continue

        # Duplicate row
        if rid in tp_ids:
            fp += 1
            continue

        gt_row = gt_map.get(rid)
        if gt_row is None:
            fp += 1  # Extra id
            continue

        # Compare all fields with GT (order doesn't matter)
        if row == gt_row:
            tp_ids.add(rid)
        else:
            fp += 1  # Field values do not match

    fn = len(gt_map) - len(tp_ids)
    precision = len(tp_ids) / (len(tp_ids) + fp) if tp_ids or fp else 0.0
    recall = len(tp_ids) / (len(tp_ids) + fn) if tp_ids or fn else 0.0
    if precision + recall == 0:
        return 0.0
    return 2 * precision * recall / (precision + recall)
\end{lstlisting}
\end{tcolorbox}    
\end{table*}

\begin{table*}
\begin{tcolorbox}[
    colframe=myblue1, 
    colback=mygray1, 
    coltitle=white, 
    title=De-duplication Task Dataset Samples, 
    % sharp corners=southwest, 
    arc=2mm, 
    boxrule=0.8mm,
    width=\textwidth
]
\centering
\begin{tabular}{|p{0.45\textwidth}|p{0.45\textwidth}|}
\hline
\textbf{Raw Data} & \textbf{GT Data} \\
\hline
% Left Column Data
\textbf{File1:}
\{ "id": "C0061", "updated\_at": "2025-04-20T13:59:30Z", "name": "Isaac", "tier": "gold" \} \newline
\{ "id": "C0024", "updated\_at": "2024-07-10T13:21:47Z", "name": "Xavier", "tier": "bronze" \} \newline
\{ "id": "C0094", "updated\_at": "2025-12-07T09:03:25Z", "name": "Queen", "tier": "gold" \} \newline
\{ "id": "C0094", "updated\_at": "2025-12-07T09:03:25Z", "name": "Queen", "tier": "gold" \} \newline
\{ "id": "C0075", "updated\_at": "2025-07-27T08:12:05Z", "name": "Xander", "tier": "bronze" \}...\newline
\textbf{File2:}
id,updated\_at,name,tier \newline
C0068,2025-06-25T00:05:48Z,Paula,silver \newline
C0107,2025-08-06T05:37:13Z,New107,silver \newline
C0072,2025-07-24T11:00:49Z,Una,gold \newline
C0062,2025-05-27T05:43:16Z,Jane,silver \newline
C0018,2024-07-21T07:27:37Z,Rupert,gold... & 
% Right Column Data
\{ "id": "C0001", "updated\_at": "2024-01-15T10:30:00Z", "name": "Alice", "tier": "gold" \} \newline
\{ "id": "C0002", "updated\_at": "2024-02-03T08:14:12Z", "name": "Bob", "tier": "silver" \} \newline
\{ "id": "C0003", "updated\_at": "2024-02-27T19:22:05Z", "name": "Carol", "tier": "bronze" \} \newline
\{ "id": "C0004", "updated\_at": "2024-03-10T07:45:51Z", "name": "Dave", "tier": "gold" \} \newline
\{ "id": "C0005", "updated\_at": "2024-03-19T11:26:31Z", "name": "Eve", "tier": "silver" \} \newline
\{ "id": "C0006", "updated\_at": "2024-03-27T15:02:43Z", "name": "Frank", "tier": "bronze" \} \newline
\{ "id": "C0007", "updated\_at": "2024-04-02T09:56:17Z", "name": "Grace", "tier": "gold" \} \newline
\{ "id": "C0008", "updated\_at": "2024-04-11T20:11:00Z", "name": "Heidi", "tier": "silver" \} \newline
\{ "id": "C0009", "updated\_at": "2024-04-23T05:33:29Z", "name": "Ivan", "tier": "bronze" \} \newline
\{ "id": "C0010", "updated\_at": "2024-04-30T18:44:07Z", "name": "Judy", "tier": "gold" \}... \\
\hline
\end{tabular}
\end{tcolorbox}    
\end{table*}

% \textbf{5. Integration Task}

% Create a tcolorbox with title and content
\begin{table*}
\begin{tcolorbox}[colframe=myblue1, colback=mygray1, coltitle=white, title=Integration Task Objective, arc=1mm, boxrule=0.8mm]
A data governance operator for composite key join: join by multi-column composite keys and resolve column conflicts. Input: customer1.csv, customer2.csv. Rule: Composite key: left(k1,k2,\dots) = right(k1',k2',\dots) (same number of columns). Conflict resolution: left-priority/right-priority/left and right suffix. Output: gt.csv.
\end{tcolorbox}
\end{table*}

\begin{table*}
\begin{tcolorbox}[colframe=myblue1, colback=mygray1, coltitle=white, title=Integration Task Eval Code, 
                 arc=1mm, boxrule=0.8mm, breakable]
\begin{lstlisting}[language=Python]
def evaluate(gt_hdr: List[str],
             gt_rows: List[Dict[str, str]],
             pred_rows: List[Dict[str, str]]) -> float:
    # 1. Column completeness
    if not pred_rows:
        print("[eval] Output is empty", file=sys.stderr)
        return 0.0

    missing = [c for c in gt_hdr if c not in pred_rows[0]]
    if missing:
        print(f"[eval] Missing columns: {missing}", file=sys.stderr)
        return 0.0

    # 2. Set comparison
    gt_counter   = rows_to_counter(gt_rows,   gt_hdr)
    pred_counter = rows_to_counter(pred_rows, gt_hdr)

    if gt_counter != pred_counter:
        lack  = gt_counter - pred_counter
        extra = pred_counter - gt_counter
        if lack:
            print(f"[eval] Missing row examples: {list(lack.elements())[:3]} ...",  file=sys.stderr)
        if extra:
            print(f"[eval] Extra row examples: {list(extra.elements())[:3]} ...", file=sys.stderr)
        return 0.0

    return 1.0
\end{lstlisting}
\end{tcolorbox}
\end{table*}

\begin{table*}
\begin{tcolorbox}[
    colframe=myblue1, 
    colback=mygray1, 
    coltitle=white, 
    title=Integration Task Dataset Samples, 
    % sharp corners=southwest, 
    arc=2mm, 
    boxrule=0.8mm,
    width=\textwidth
]
\centering
\begin{tabular}{|p{0.45\textwidth}|p{0.45\textwidth}|}
\hline
\textbf{Raw Data} & \textbf{GT Data} \\
\hline
% Left Column Data
\textbf{File1:} \newline
country,region,customer\_id,email, \newline signup\_date,status,notes \newline
US,CA,1001,alice@example.com,2021-01-10,active,L1 \newline
US,NY,1002,bob@example.com,2021-02-12,inactive,L2 \newline
CN,BJ,2001,chen@example.cn,2020-11-05,active,L3 \newline
CN,SH,2002,du@example.cn,2022-07-19,pending,L4 \newline
DE,BE,3001,eva@example.de,2021-09-30,active,L5 \newline
US,CA,1003,frank@example.com,2020-06-15,active,L6 \newline
\textbf{File2:} \newline
country\_code,region,id,email, \newline last\_order\_date,status,vip \newline
US,CA,1001,alice.us@example.com,\newline 2022-12-01,gold,true \newline
US,NY,1002,bob@example.com,2021-12-11,inactive,false \newline
CN,BJ,2001,chen\_new@ex.cn,2023-03-03,active,true \newline
CN,GD,2005,gao@example.cn,2021-05-05,active,false \newline
DE,BE,3001,eva@example.de,\newline 2022-02-02,paused,false \newline
US,CA,9999,zoe@example.com,2023-04-04,active,false \newline
US,CA,1003,frank@example.com,2020-07-01,inactive,false \newline
CN,SH,2002,du@alt.cn,2022-08-01,active,true & 
% Right Column Data
country,region,customer\_id,email\_left, \newline signup\_date,status\_left,notes,email\_right \newline last\_order\_date,status\_right,vip \newline
US,CA,1001,alice@example.com,2021-01-10,active,L1,alice.us@example.com,\newline 2022-12-01,gold,true \newline
US,NY,1002,bob@example.com,2021-02-12,inactive,L2,bob@example.com, \newline 2021-12-11,inactive,false \newline
CN,BJ,2001,chen@example.cn,2020-11-05,active,L3,chen\_new@ex.cn,2023-03-03,active,true \newline
CN,SH,2002,du@example.cn,2022-07-19,pending,L4,du@alt.cn,2022-08-01,active,true \newline
DE,BE,3001,eva@example.de,2021-09-30,active,L5,eva@example.de,2022-02-02,paused,false \newline
US,CA,1003,frank@example.com,2020-06-15,active,L6,frank@example.com, \newline 2020-07-01,inactive,false \\
\hline
\end{tabular}
\end{tcolorbox}
\end{table*}

% \textbf{6. Classification and Labeling Task}

% Create a tcolorbox with title and content
\begin{table*}
\begin{tcolorbox}[colframe=myblue1, colback=mygray1, coltitle=white, title=Classification and Labeling Task Objective, arc=1mm, boxrule=0.8mm]
Use LLMserving to assign sentiment labels to text: Input format: \texttt{.jsonl} with \texttt{text\_id} and \texttt{content}; Sentiment label set: Positive / Neutral / Negative.
\end{tcolorbox}
\end{table*}

\begin{table*}
\begin{tcolorbox}[colframe=myblue1, colback=mygray1, coltitle=white, title=Classification and Labeling Task Eval Code, 
                 arc=1mm, boxrule=0.8mm, breakable]
\begin{lstlisting}[language=Python]
def accuracy(gt: List[Dict[str, Any]], pred: List[Dict[str, Any]]) -> float:
    """
    Calculate the simple classification accuracy between predictions and ground truth.
    """
    # Create {text_id: sentiment} mapping; trim leading and trailing spaces and standardize case
    norm = lambda s: str(s).strip()  # Only trim; case-sensitive
    gt_map   = {norm(r["text_id"]): norm(r["sentiment"]) for r in gt}
    pred_map = {norm(r["text_id"]): norm(r.get("sentiment", "")) for r in pred}

    total   = len(gt_map)
    correct = sum(1 for k, v in gt_map.items() if pred_map.get(k) == v)
    return correct / total if total else 0.0
\end{lstlisting}
\end{tcolorbox}
\end{table*}

\begin{table*}
\begin{tcolorbox}[
    colframe=myblue1, 
    colback=mygray1, 
    coltitle=white, 
    title=Classification and Labeling Task Dataset Samples, 
    % sharp corners=southwest, 
    arc=2mm, 
    boxrule=0.8mm,
    width=\textwidth
]
\centering
\begin{tabular}{|p{0.45\textwidth}|p{0.45\textwidth}|}
\hline
\textbf{Raw Data} & \textbf{GT Data} \\
\hline
% Left Column Data
\{"text\_id": "0001", "content": "The latte at this coffee shop is so delicious, I will definitely come back next time!"\} \newline
\{"text\_id": "0002", "content": "The customer service response speed is quite fast, and the problem has been solved."\} \newline
\{"text\_id": "0003", "content": "The sunlight today is really nice, feeling great."\} \newline
\{"text\_id": "0004", "content": "The soundtrack of this movie is very moving, definitely recommend it."\} \newline
\{"text\_id": "0005", "content": "The project was launched on time, and everyone is very satisfied."\} \newline
\{"text\_id": "0079", "content": "This is the second page of the contract."\} \newline
\{"text\_id": "0080", "content": "The air conditioning temperature is set to 25°C."\} \newline
\{"text\_id": "0081", "content": "The service attitude was terrible, I will never come again."\} \newline
\{"text\_id": "0082", "content": "The product broke after just two days of use, very disappointing."\} \newline
\{"text\_id": "0083", "content": "The courier hasn't updated the logistics for a week, so annoying."\} &
% Right Column Data
\{"text\_id":"0001","content":"The latte at this coffee shop is so delicious, I will definitely come back next time!","sentiment":"Positive"\} \newline
\{"text\_id":"0002","content":"The customer service response speed is quite fast, and the problem has been solved.","sentiment":"Positive"\} \newline
\{"text\_id":"0003","content":"The sunlight today is really nice, feeling great.","sentiment":"Positive"\} \newline
\{"text\_id":"0004","content":"The soundtrack of this movie is very moving, definitely recommend it.","sentiment":"Positive"\} \newline
\{"text\_id":"0005","content":"The project was launched on time, and everyone is very satisfied.","sentiment":"Positive"\} \newline
\{"text\_id":"0079","content":"This is the second page of the contract.","sentiment":"Neutral"\} \newline
\{"text\_id":"0080","content":"The air conditioning temperature is set to 25°C.","sentiment":"Neutral"\} \newline
\{"text\_id":"0081","content":"The service attitude was terrible, I will never come again.","sentiment":"Negative"\} \newline
\{"text\_id":"0082","content":"The product broke after just two days of use, very disappointing.","sentiment":"Negative"\} \newline
\{"text\_id":"0083","content":"The courier hasn't updated the logistics for a week, so annoying.","sentiment":"Negative"\} \\
\hline
\end{tabular}
\end{tcolorbox}
\end{table*}

\newpage
% \textbf{7. DAG Task}

% Create a tcolorbox with title and content
\begin{table*}
\begin{tcolorbox}[colframe=myblue1, colback=mygray1, coltitle=white, title=DAG Task Objective, arc=1mm, boxrule=0.8mm]
Write an operator to process JSONL files, executing sequentially: filter out records with a high proportion of symbols in the text field $\rightarrow$ remove excess spaces in the text field $\rightarrow$ censor profanity in the text field with ****, for example, ``I am fucking happy'' becomes ``I am **** happy'' $\rightarrow$ use MinHash for approximate deduplication ($\geq$0.9), retaining the record with the smallest id; output JSONL.
\end{tcolorbox}
\end{table*}

\begin{table*}
\begin{tcolorbox}[colframe=myblue1, colback=mygray1, coltitle=white, title=DAG Task Eval Code, 
                 arc=1mm, boxrule=0.8mm, breakable]
\begin{lstlisting}[language=Python]
def evaluate(processed_path):
    expected_path = get_gt()
    expected = load_jsonl(expected_path)
    processed = load_jsonl(processed_path)

    # Construct mappings for comparison
    expected_map = {entry["id"]: entry for entry in expected}
    processed_map = {entry["id"]: entry for entry in processed}

    # Only evaluate the intersection part
    common_ids = set(expected_map.keys()) & set(processed_map.keys())

    true_positives = 0
    for cid in common_ids:
        gt = expected_map[cid]
        pred = processed_map[cid]

        # Check if text is the same (strip leading and trailing spaces)
        if gt["text"].strip() == pred["text"].strip():
            true_positives += 1

    predicted_total = len(processed_map)
    gold_total = len(expected_map)

    precision = true_positives / predicted_total if predicted_total > 0 else 0.0
    recall = true_positives / gold_total if gold_total > 0 else 0.0
    f1 = (2 * precision * recall) / (precision + recall) if precision + recall > 0 else 0.0

    result = {"eval_score": f"{f1:.4f}"}
    print(result)
\end{lstlisting}
\end{tcolorbox}
\end{table*}

\begin{table*}
\begin{tcolorbox}[
    colframe=myblue1, 
    colback=mygray1, 
    coltitle=white, 
    title=DAG Task Dataset Samples, 
    % sharp corners=southwest, 
    arc=2mm, 
    boxrule=0.8mm,
    width=\textwidth,
    breakable
]
\centering
\begin{tabular}{|p{0.45\textwidth}|p{0.45\textwidth}|}
\hline
\textbf{Raw Data} & \textbf{GT Data} \\
\hline
% Left Column Data
\{"id": 1, "items": ["orange", "computer", "paper", "pear", "book", "phone"], "text": "Sports and the environment have a complex relationship that requires careful consideration and action if we want to keep enjoying both. On one hand, sporting events bring people together, promote health, and drive the economy. On the other hand, they can be a asshole environmental nightmare", "sources": ["dataset\_b.jsonl", "dataset\_a.jsonl"]\} \newline
\{"id": 26, "items": ["orange", "book", "banana", "grape", "computer"], "text": "Engaging in sports is one hell of a way to boost your overall health and well-being, both physically and mentally. Whether you're hitting the gym, playing soccer, or going for a run, these activities keep your", "sources": ["dataset\_b.jsonl", "dataset\_c.jsonl"]\} \newline
\{"id": "d4a6cae8-6250-40dc-9a1e-b9bef91620fd", "items": ["pen", "orange", "grape", "computer", "banana", "paper"], "text": "Art has a **** magical way of weaving itself into the fabric of health, providing both mental clarity and emotional solace. Through the **** strokes of a paintbrush ???????????????????????????????????
\newline
???????????????????????????????????\newline
??????????????????????????????????? or the rhythmic beats of a song, art offers a therapeutic escape from life's", "sources": ["dataset\_b.jsonl"]\} &
% Right Column Data
\{"id": 1, "items": ["orange", "computer", "paper", "pear", "book", "phone"], "text": "Sports and the environment have a complex relationship that requires careful consideration and action if we want to keep enjoying both. On one hand, sporting events bring people together, promote health, and drive the economy. On the other hand, they can be a **** environmental nightmare", "sources": ["dataset\_b.jsonl", "dataset\_a.jsonl"]\} \newline
\{"id": 26, "items": ["orange", "book", "banana", "grape", "computer"], "text": "Engaging in sports is one hell of a way to boost your overall health and well-being, both physically and mentally. Whether you're hitting the gym, playing soccer, or going for a run, these activities keep your", "sources": ["dataset\_b.jsonl", "dataset\_c.jsonl"]\} \\
\hline
\end{tabular}
\end{tcolorbox}
\end{table*}

\newpage
\subsection{Agent roles and implementation details}
\label{appendix:agent roles}

\textbf{The Planner: From Intent to High-Level DAG. }
The initial phase is dedicated to understanding the user's goal and formulating a strategic plan. This is achieved through two sequential tasks:

\begin{itemize}
\item \textit{\textbf{Intent Understanding:}} Upon receiving a natural language request, the Planner leverages a LLM configured with \textbf{few-shot prompting}. It analyzes the user's intent by conditioning the model with the provided data schema and data samples. This grounding process ensures the user's goal is not only correctly interpreted but also validated for feasibility against the actual data context.
\item \textit{\textbf{Contract-Guided Planning:}} After intent understanding, the Planner does not directly generate a concrete blueprint. Instead, it first extracts verifiable governance contracts from the user request, data schema, and data samples. Each contract is attached to an operator in the form of a 2-tuple $(\textsc{pre}, \textsc{post})$, strictly defining the pre-conditions and post-conditions for execution. The Planner then generates a sequence that satisfies the constraints imposed by these contracts, ensuring that the output ($\textsc{post}$) of each step fulfills the input requirements ($\textsc{pre}$) of the subsequent step. When a constraint is not met, the system automatically inserts minimal repair steps (such as imputation or type casting).
\item \textit{\textbf{Pipeline Recommendation:}} Building on the above deep understanding and contract-guided planning, the Planner ultimately formulates a high-level governance plan, which is represented as a preliminary directed acyclic graph (DAG). The nodes of this DAG correspond to a series of abstract operators (e.g., ``Remove Duplicates", ``Standardize Date Format", ``Impute Missing Values"). These contract-annotated nodes collectively provide a strategic blueprint for the subsequent execution phase, ensuring that the final generated code strictly adheres to the validated logical path.
\end{itemize}

\textbf{The Executor: Realizing Operators with Retrieval-Augmented Generation. }For each abstract operator in the planned DAG, the Executor is responsible for generating concrete, executable Python code. It employs a powerful \textbf{Retrieval-Augmented Generation (RAG)} strategy, which synergizes the reliability of pre-validated code with the flexibility of on-the-fly generation.

\begin{itemize}
\item \textit{\textbf{Operator Retrieval:}} The agent first treats its internal library of validated governance operators as a collection of callable \textbf{tools}. Each tool has a rich description detailing its functionality, parameters, and use cases. The Executor compares the semantic content of the target operator's goal (e.g., ``standardize date format to YYYY-MM-DD") against these tool descriptions to retrieve the top-K (e.g., top-4) most relevant operators.
\item \textit{\textbf{Augmented Generation:}} Rather than simply executing the top retrieved operator or falling back to free generation if no perfect match is found, the Executor adopts a more robust approach. The retrieved operators, along with their descriptions, are injected into the LLM's prompt as dynamic few-shot examples. This context-rich prompt guides the model to generate code that is not only tailored to the specific requirements of the task but also adheres to the established patterns and best practices of the operator library. This hybrid method significantly reduces hallucinations and improves the quality of the generated code, even for highly customized or novel tasks.
\end{itemize}

\textbf{The Evaluator: Sandboxed Execution and feedback-driven debugging. }Code generation is only half the battle; rigorous verification is paramount. The Evaluator provides a critical quality assurance layer through a self-correcting execution and debugging cycle.

\begin{itemize}
\item \textit{\textbf{Sandboxed Execution:}} All generated code is executed within a secure, isolated sandbox environment. This prevents unintended side effects on the host system and allows the agent to safely handle diverse data sources and external dependencies.
\item \textit{\textbf{Iterative Debugging with Structured Feedback:}} When the generated code fails to execute or produces incorrect results, the Evaluator does not simply report the failure. Instead, it acts as a diagnostician, capturing the runtime state and constructing a highly structured feedback prompt to guide the Executor’s subsequent refinement. As shown in Figure \ref{fig:architecture}, this prompt is a rich data object containing a comprehensive diagnostic report: it includes not only the erroneous code snippet that caused the failure, but also the complete error message and stack trace, providing technical context for issue localization. More importantly, the Evaluator also analyzes the situation in light of the relevant \textbf{contract constraints}. If any contract is found to be unsatisfied, it offers targeted revision suggestions—for example, \textit{Please add a check to handle potential null values in the creation\_date column before applying the datetime conversion.”} To keep the agent aligned with the overall objective, the feedback additionally includes broader task context. \end{itemize} 

This feedback allows the Executor to perform targeted, specific fixes instead of trial-and-error guessing. This loop continues until the operator code is both runnable and functionally correct, ensuring each component of the final GovDAG is rigorously validated.

\subsection{Details of Agent Framework Baselines}
\label{appendix:details}
\subsubsection{Derived Metrics and Formulas}
The following metrics are used to evaluate agent performance throughout the appendix.
\begin{itemize}
\item \textbf{Alignment:} $A = \text{TSR} / \text{CRR}$.
\item \textbf{Contract gap:} $\Delta_{rc} = \text{CRR} - \text{TSR}$ (in percentage points).
\item \textbf{Debugging efficiency:} $E = \text{TSR} / \text{ADI}$.
\item \textbf{Tokens per successful task:} $T^* = \text{Avg. Tokens} / (\text{TSR}/100)$. This measures the average number of tokens consumed to achieve one successful task completion.
\end{itemize}

\begin{figure*}[h!]
    \centering
    \includegraphics[width=\textwidth]{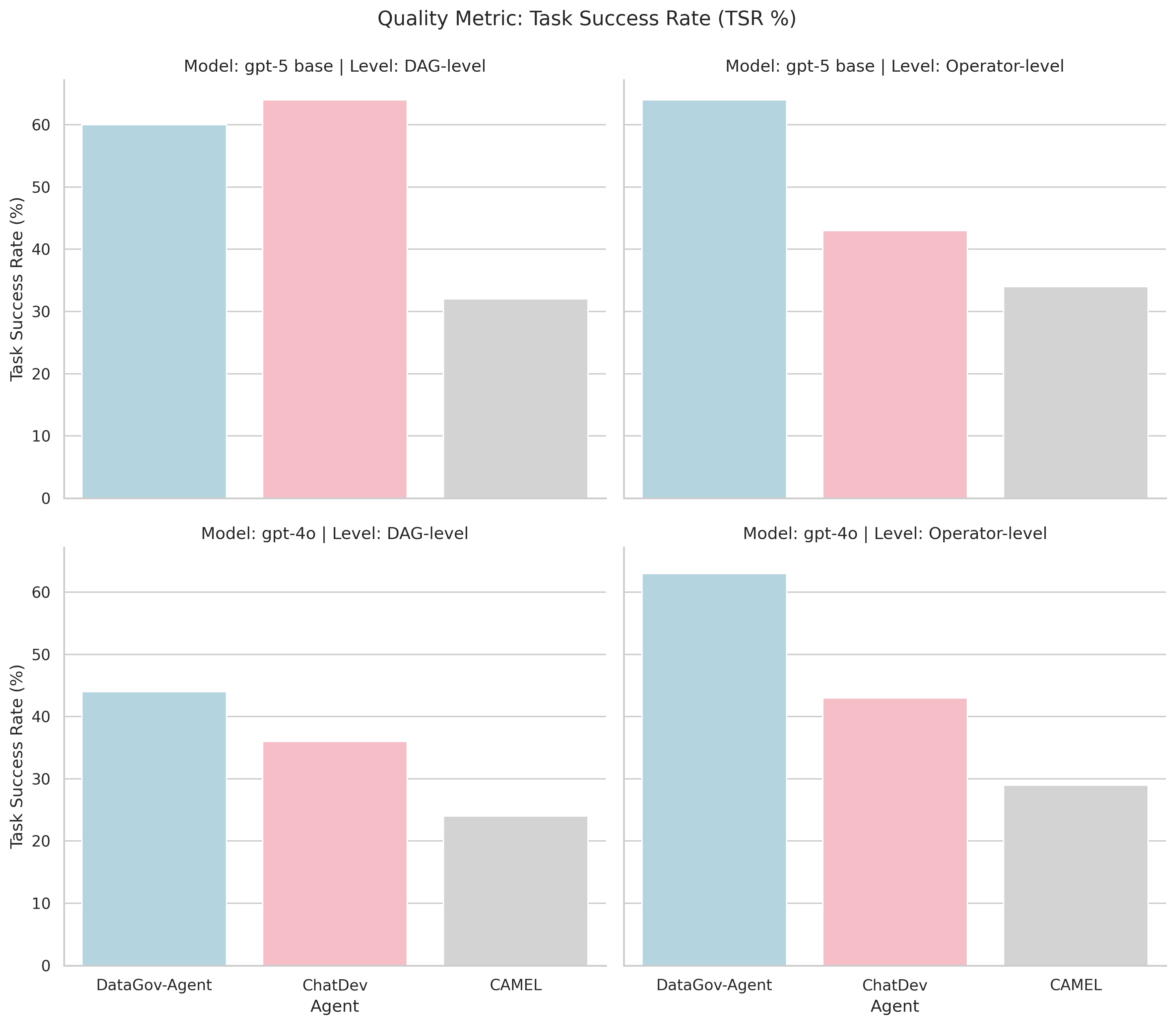}
    \caption{Comparison of Task Success Rate (TSR) across agents, base models, and task levels. TSR measures the percentage of tasks completed successfully.}
    \label{fig:appendix_tsr}
\end{figure*}

\begin{figure*}[h!]
    \centering
    \includegraphics[width=\textwidth]{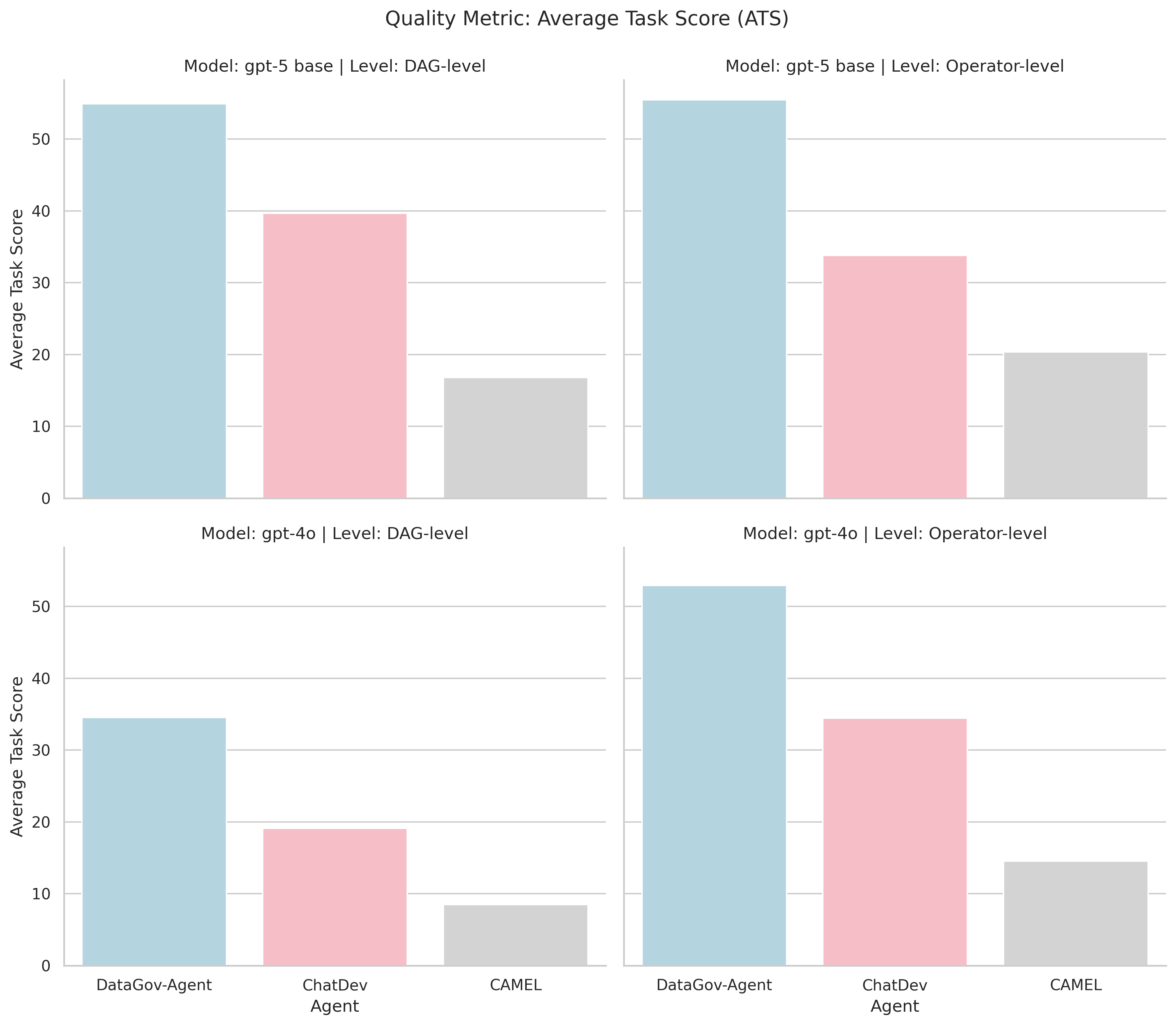}
    \caption{Comparison of Average Task Score (ATS). ATS provides a more nuanced measure of solution quality beyond simple success or failure.}
    \label{fig:appendix_ats}
\end{figure*}

\begin{figure*}[h!]
    \centering
    \includegraphics[width=\textwidth]{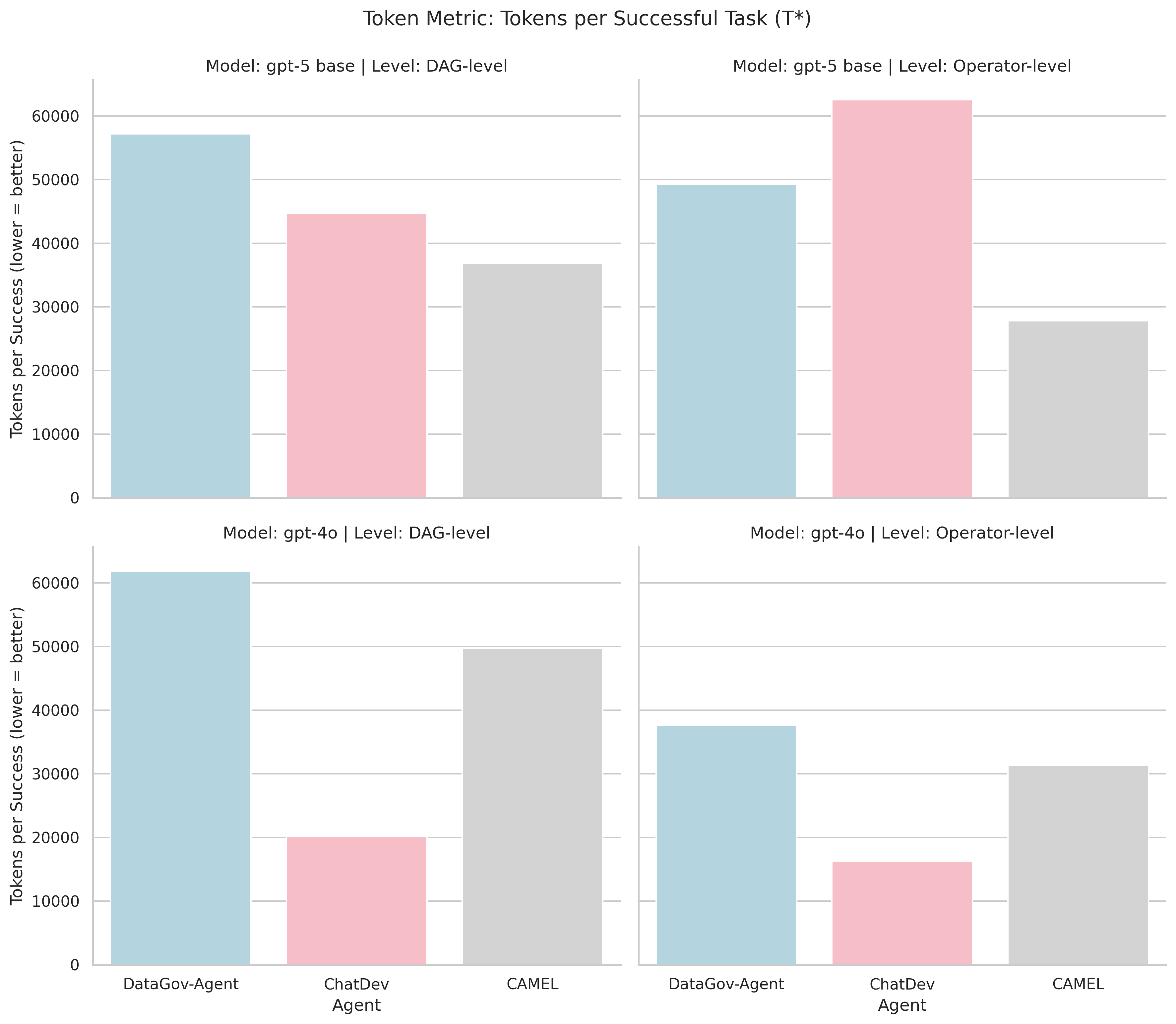}
    \caption{Comparison of Tokens per Successful Task ($T^{*}$). This metric normalizes average token consumption by the success rate, indicating token-efficiency. Lower values are better.}
    \label{fig:appendix_t_star}
\end{figure*}

\begin{figure*}[h!]
    \centering
    \includegraphics[width=\textwidth]{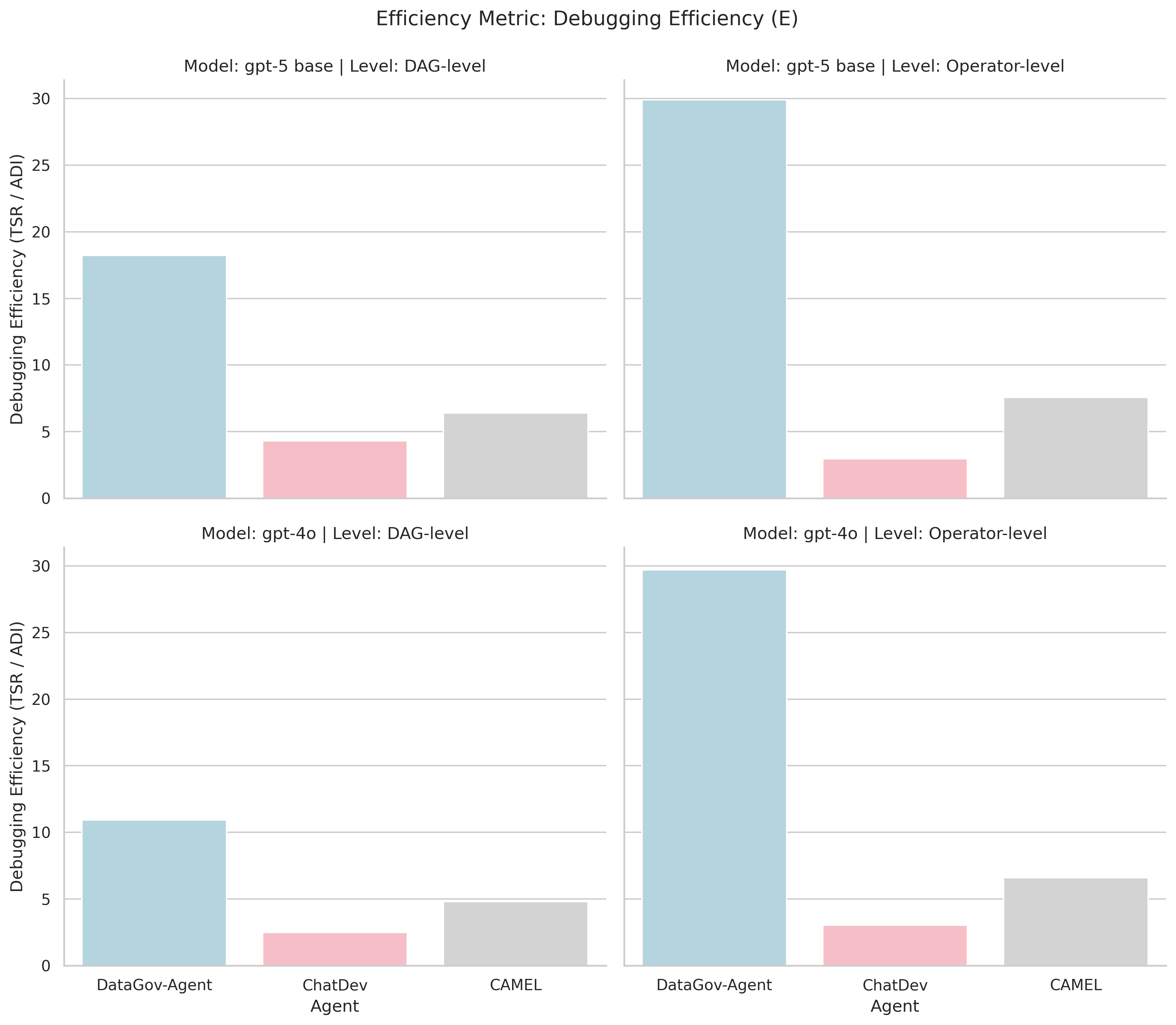}
    \caption{Comparison of Debugging Efficiency ($E$). This metric reflects how many successful tasks are produced per debugging iteration. Higher values are better.}
    \label{fig:appendix_efficiency}
\end{figure*}

\begin{figure*}[h!]
    \centering
    \includegraphics[width=\textwidth]{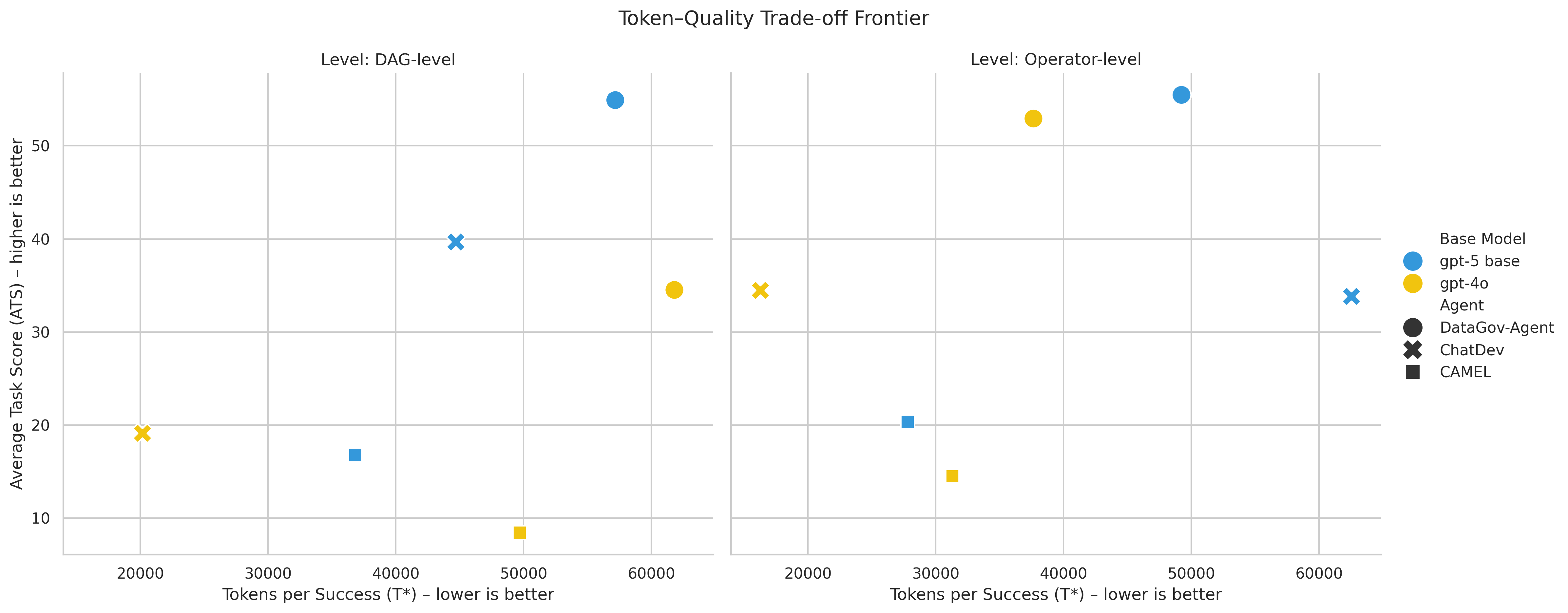}
    \caption{Token–Quality Trade-off Frontier. Relationship between quality (ATS, $y$-axis) and token efficiency ($T^{*}$, $x$-axis). The ideal position is the top-left corner (high quality, low tokens per success).}
    \label{fig:appendix_tradeoff}
\end{figure*}

The following sections provide the specific numerical data and interpretations corresponding to the visualizations in Figures \ref{fig:appendix_tsr} through \ref{fig:appendix_tradeoff}.

\paragraph{GPT-5 base -- DAG-level details}
\begin{itemize}
\item \textbf{DataGovAgent} (TSR 60, CRR 74, ATS 54.91, Avg. 62.97, ADI 3.29, Tokens 34303.72) \
$A = 0.81$; $\Delta_{rc} = 14$; $E = 18.24$; $T^* = 57,173$.
\item \textbf{ChatDev} (64, 82, 39.67, 61.89, 14.89, Tokens 28607.22) \\
$A = 0.78$; $\Delta_{rc} = 18$; $E = 4.30$; $T^* = 44,700$.

\item \textbf{CAMEL} (32, 74, 16.80, 40.93, 5.00, Tokens 11777.50) \\
$A = 0.43$; $\Delta_{rc} = 42$; $E = 6.40$; $T^* = 36,805$.
\end{itemize}
\textit{Interpretation: On complex DAG-level tasks, DataGovAgent demonstrates the highest debugging efficiency (E=18.24) and strong alignment (A=0.81). However, this comes at the highest token cost per successful task ($T^{*}=57,173$). In contrast, CAMEL is the most token-efficient per success ($T=36,805$) but delivers significantly lower quality (TSR 32, ATS 16.80) and poor alignment. ChatDev offers a middle ground on token efficiency but lags considerably in debugging efficiency.}

\paragraph{GPT-5 base -- Operator-level details}
\begin{itemize}
\item \textbf{DataGovAgent} (TSR 64, CRR 88, ATS 55.47, Avg. 69.15, ADI 2.14, Tokens 31503.75) \
$A = 0.73$; $\Delta_{rc} = 24$; $E = 29.91$; $T^* = 49,225$.
\item \textbf{ChatDev} (43, 69, 33.82, 48.61, 14.47, Tokens 26888.26) \\
$A = 0.62$; $\Delta_{rc} = 26$; $E = 2.97$; $T^* = 62,531$.

\item \textbf{CAMEL} (34, 92, 20.36, 48.79, 4.50, Tokens 9447.75) \\
$A = 0.37$; $\Delta_{rc} = 58$; $E = 7.56$; $T^* = 27,788$.
\end{itemize}
\textit{Interpretation: Even on simpler Op-level tasks, DataGovAgent leads in quality (TSR 64, ATS 55.47) and debugging efficiency (E=29.91). It is also more token-efficient per success than ChatDev ($T=49,225$ vs. $62,531$). CAMEL remains the most token-efficient overall ($T=27,788$) but has the worst alignment (A=0.37) and a large correctness gap ($\Delta_{rc}=58$), indicating that while its raw token usage is low, it struggles to convert runnability into correct solutions.}
\paragraph{Weaker base model (GPT-4o) -- token-quality trade-off}
\textbf{DAG-level:}
\begin{itemize}
\item \textbf{DataGovAgent} (44, 50, 34.52, 42.84, 4.03, Tokens 27192.45): $A=0.88$; $\Delta_{rc}=6$; $E=10.92$; $T^{*}=61,801$.
\item \textbf{ChatDev} (36, 40, 19.12, 31.71, 14.42, Tokens 7261.49): $A=0.90$; $\Delta_{rc}=4$; $E=2.50$; $T^{*}=20,171$.
\item \textbf{CAMEL} (24, 60, 8.47, 30.82, 5.00, Tokens 11925.00): $A=0.40$; $\Delta_{rc}=36$; $E=4.80$; $T^*=49,688$.
\end{itemize}

\textbf{Operator-level:}
\begin{itemize}
\item \textbf{DataGovAgent} (63, 89, 52.93, 68.31, 2.12, Tokens 23712.14): $A=0.71$; $\Delta_{rc}=26$; $E=29.72$; $T^{*}=37,638$.
\item \textbf{ChatDev} (43, 63, 34.47, 46.82, 14.20, Tokens 6996.62): $A=0.68$; $\Delta_{rc}=20$; $E=3.03$; $T^{*}=16,271$.
\item \textbf{CAMEL} (29, 91, 14.54, 44.85, 4.40, Tokens 9071.92): $A=0.32$; $\Delta_{rc}=62$; $E=6.59$; $T^{*}=31,282$.
\end{itemize}
\textit{Interpretation: With the weaker GPT-4o model, the trade-offs become more pronounced. DataGovAgent still achieves the highest quality (TSR/ATS) and debugging efficiency (E), but at a significantly higher token cost per success ($T^{*}$). Surprisingly, ChatDev becomes the most token-efficient framework ($T*$ of 20,171 on DAG and 16,271 on Op), despite its low raw success rate and poor debugging efficiency. This highlights a clear, controllable token-quality frontier where achieving higher quality and development efficiency with DataGovAgent requires a larger token budget.}

\subsubsection{Performance Visualizations}
The following figures provide a comparative visualization of agent performance across different models, task levels, and key metrics.

 % Prevents text from awkwardly wrapping around large figures

\subsubsection{Mechanism Analysis and Ablation Study}
\textbf{Contracts enforce executable correctness.} Pre-conditions explicate assumptions regarding data types, shapes, and constraints (e.g., uniqueness), while post-conditions translate acceptance criteria into assertions, preventing silent failures across steps.

\textbf{Feedback-driven debugging resolves execution failures.} The Evaluator generates failing code spans, stack traces, and violated contracts to generate targeted repairs. This mechanism improves accuracy ($A$) and reduces redundancy ($\Delta_{rc}$) with significantly higher efficiency ($E$).

\textbf{Ablation Results.} Table \ref{tab:ablation_study} presents the impact of removing key components (using GPT-5 base):
\begin{itemize}
\item \textbf{w/o Planner:} Performance degrades significantly, with TSR dropping from $64\%$ to $38\%$ ($-26$ pp) and CRR falling from $88\%$ to $51\%$. ADI worsens ($2.14 \rightarrow 8.75$), indicating a loss of structural coherence.
\item \textbf{w/o RAG:} TSR declines to $49\%$ ($-15$ pp) and CRR to $65\%$, confirming that retrieval is essential for reducing hallucinations.
\end{itemize}
These results confirm that contract-guided planning ensures correct task decomposition, while the Evaluator's iterative loop converts these foundations into efficient code generation.

\subsection{Ablation Study}
% \section{Ablation Study}
\label{sec:ablation}

To dissect the contribution of each component within the \textsc{DataGovAgent} framework, we conducted a series of ablation studies on the DataGovbench Operator-level tasks. We systematically disabled or replaced key modules—the Planner and the RAG mechanism to quantify their impact on overall performance. All experiments were run using GPT-5 as the base model. The results are summarized in Table \ref{tab:ablation_study}.

\begin{table}[htbp]
  \centering
  \caption{Ablation study of DataGovAgent on DataGovbench operator-level tasks. %
           Numbers in brackets show the change (∆) w.r.t.\ the full model
           — red = decrease, green = increase.}
  \label{tab:ablation_study}
  \renewcommand{\arraystretch}{1.2}

  % 自动按列宽缩放整个表格（含字号）
  \resizebox{\columnwidth}{!}{%
    \begin{tabular}{
        l                       % Configuration
        S[table-format=2.2]     % ATS
        S[table-format=2.2]     % TSR
        S[table-format=2.2]     % CRR
        S[table-format=2.2]     % ADI
    }
      \toprule
      \textbf{Configuration} &
      {\textbf{ATS$\uparrow$}} & {\textbf{TSR$\uparrow$}} &
      {\textbf{CRR$\uparrow$}} & {\textbf{ADI$\downarrow$}} \\
      \midrule
      \rowcolor{gray!10}
      \textbf{DataGovAgent (Full)} &
        \textbf{55.47} & \textbf{64.00} & \textbf{88.00} & \textbf{2.14} \\
      \midrule
      \multicolumn{5}{l}{\textit{RQ1: Planner's Role}} \\[-0.25em]
      \quad w/o Planner &
        {31.20 (\textcolor{red}{-24.27})} &
        {38.00 (\textcolor{red}{-26.00})} &
        {51.00 (\textcolor{red}{-37.00})} &
        {8.75  (\textcolor{green}{+6.61})} \\
      \midrule
      \multicolumn{5}{l}{\textit{RQ2: RAG's Impact}} \\[-0.25em]
      \quad w/o RAG (Free Generation) &
        {42.15 (\textcolor{red}{-13.32})} &
        {49.00 (\textcolor{red}{-15.00})} &
        {65.00 (\textcolor{red}{-23.00})} &
        {5.20  (\textcolor{green}{+3.06})} \\
      \bottomrule
    \end{tabular}%
  }
\end{table}

% \begin{table*}[htbp]
%   \centering
%   \caption{Ablation study of DataGovAgent on DataGovbench operator-level tasks. %
%            Numbers in brackets show the change (∆) w.r.t.\ the full model
%            — red = decrease, green = increase.}
%   \label{tab:ablation_study}
%   \renewcommand{\arraystretch}{1.2}

%   %------------------------------------------------------
%   \begin{tabular}{
%       l                       % Configuration
%       S[table-format=2.2]     % ATS
%       S[table-format=2.2]     % TSR
%       S[table-format=2.2]     % CRR
%       S[table-format=2.2]     % ADI
%   }
%     \toprule
%     \textbf{Configuration} &
%     {\textbf{ATS$\uparrow$}} & {\textbf{TSR$\uparrow$}} &
%     {\textbf{CRR$\uparrow$}} & {\textbf{ADI$\downarrow$}} \\
%     \midrule
%     \rowcolor{gray!10}
%     \textbf{DataGovAgent (Full)} &
%       \textbf{55.47} & \textbf{64.00} & \textbf{88.00} & \textbf{2.14} \\
%     \midrule
%     \multicolumn{5}{l}{\textit{RQ1: Planner's Role}} \\[-0.25em]
%     \quad w/o Planner &
%       {31.20 (\textcolor{red}{-24.27})} &
%       {38.00 (\textcolor{red}{-26.00})} &
%       {51.00 (\textcolor{red}{-37.00})} &
%       {8.75  (\textcolor{green}{+6.61})} \\
%     \midrule
%     \multicolumn{5}{l}{\textit{RQ2: RAG's Impact}} \\[-0.25em]
%     \quad w/o RAG (Free Generation) &
%       {42.15 (\textcolor{red}{-13.32})} &
%       {49.00 (\textcolor{red}{-15.00})} &
%       {65.00 (\textcolor{red}{-23.00})} &
%       {5.20  (\textcolor{green}{+3.06})} \\
%     % \midrule

%     \bottomrule
%   \end{tabular}
% \end{table*}

\paragraph{RQ1: Is the Planner's high-level DAG planning necessary?}
To answer this, we created a variant named \textbf{`w/o Planner`}, where the Executor directly receives the raw natural language instruction and attempts to generate the entire solution in one go, bypassing the intent understanding and DAG planning phase. As shown in Table \ref{tab:ablation_study}, this led to a catastrophic performance drop: the TSR plummeted from 64.00\% to 38.00\%, and the Average Debug Iterations (ADI) quadrupled. This result strongly indicates that for data governance tasks, which often involve implicit multi-step logic, decomposing the user's intent into a structured, high-level plan is crucial. Without this planning phase, the LLM struggles to manage the complexity, leading to logically flawed or incomplete code that is difficult to debug.

\paragraph{RQ2: How much does Retrieval-Augmented Generation contribute?}
We investigated this by creating the \textbf{`w/o RAG'} variant, where the Executor generates code based solely on the abstract operator name provided by the Planner, without retrieving any code examples from the operator library. The performance degradation was significant, with TSR dropping by 15 percentage points. This highlights the value of RAG: grounding the LLM with pre-validated, high-quality code snippets (even if they are not a perfect match) significantly steers it towards generating more correct and robust solutions, reducing hallucinations and logical errors.

\newpage
\subsection{Metrics}
\label{appendix:a4}
See the metric details in Table~\ref{tab:evaluation_metrics_final}

\begin{table*}[htbp]
\centering
% 使用 threeparttable 环境来包裹表格和注释
\begin{threeparttable}
\caption{Evaluation Metrics for DataGovbench}
\label{tab:evaluation_metrics_final}
\begin{tabularx}{\textwidth}{@{} ll >{\centering\arraybackslash}c >{\RaggedRight}X @{}}
\toprule
\textbf{Metric} & \textbf{Abbr.} & \textbf{Calculation} & \textbf{Description} \\
\midrule

Average Task Score & ATS & 
$\dfrac{100}{N_t} \sum_{i=1}^{N_t} S_i$ & 
Represents the ATS across all tasks, reflecting the overall quality of the generated solutions. A higher ATS indicates better overall performance. \\
\addlinespace

Task Success Rate & TSR & 
$\dfrac{N_\text{succ}}{N_t}$ & 
The proportion of tasks that fully achieve the ``business objective." This is the core metric for measuring task completion quality. \\
\addlinespace

Code Runnable Rate & CRR & 
$\dfrac{N_\text{run}}{N_\text{gen}}$ & 
The proportion of generated code scripts that can be executed directly without any uncaught errors. This measures the basic usability of the code. \\
\addlinespace

Avg. Score & -- &
$S_{avg}$ &
The average value of the ATS, TSR, and CRR metrics. This metric provides an overall score by averaging these three indicators. \\
\addlinespace

Average Debug Iterations & ADI & 
$\dfrac{1}{N_t} \sum_{i=1}^{N_t} D_i$ & 
The average number of ``generate → execute → evaluate" cycles required for a task to succeed. This measures the debugging efficiency of the agent framework. \\
\addlinespace

Avg. Tokens & -- & 
$T_{avg}$ & 
The average number of tokens consumed to complete each individual task. \\
\addlinespace

Total Cost & -- & 
$C_i$ & 
The monetary cost required to complete each individual task, calculated based on openai LLM API pricing. This metric evaluates the economic efficiency for every single task. \\
\addlinespace
Generation Time & -- & 
$T_\text{gen}$ & 
Total wall-clock time (in seconds) consumed by the LLM to generate all task code solutions. This reflects the raw code synthesis efficiency. \\
\addlinespace

Execution Time & -- & 
$T_\text{exec}$ & 
Total wall-clock time (in seconds) consumed by running all generated task code solutions. This reflects the runtime efficiency of the produced code. \\

\bottomrule
\end{tabularx}

% 表格下方的注释区域
\begin{tablenotes}[para,flushleft]
    \footnotesize
    \item[] \textbf{Where:}$N_t$ is the total number of tasks; $S_i$ is the evaluation score for task $i$; $N_\text{succ}$ is the number of successful tasks; $N_\text{run}$ is the number of runnable scripts; $N_\text{gen}$ is the total number of generated scripts; $D_i$ is the number of debug iterations for task $i$; $C_{i}$ is the monetary cost for each individual task; $T_\text{gen}$ is the total generation time across all tasks; and $T_\text{exec}$ is the total execution time across all tasks. \textbf{Notes:} ATS = 100 × mean per‑task score (each task score $\in$ [0,1]). TSR/CRR are proportions reported as percentages. Higher is better unless noted.
\end{tablenotes}
\end{threeparttable}
\end{table*}

\newpage
\subsection{Prompts}
\label{appendix:Prompts}
Here's some prompt templates used in Benchmark Building.

\renewcommand\lstlistingname{Prompt}

\begin{lstlisting}[breaklines=true,label={prompt:dag},caption={Prompt for building DAG tasks.}]
### Task Description
You are given a sequence of task descriptions. Each task description defines a part of a complex task or operation. The task descriptions are part of a larger, multi-step process that will form a comprehensive, integrated task. Your objective is to generate a new, high-level task objective that combines the individual task descriptions into a coherent and complex task. This task must challenge the model's ability to handle intricate data governance problems.

### Instructions
1. Combine the given task descriptions into a single, cohesive task that requires handling multiple steps.
2. Incorporate multiple aspects of the given task descriptions into the final task description to present a significant challenge to data governance.

### Task Descriptions
- {task_1}
- {task_2}
- {task_3}
- ...

### Generated Comprehensive Task
{generated_task}
\end{lstlisting}

\renewcommand\lstlistingname{Prompt}

\begin{lstlisting}[breaklines=true,label={prompt:reverse},caption={Prompt for reverse prompt.}]
### Original Task Objective
You are given the following task objective. Your goal is to achieve the stated objective using the provided data examples.

### Task Description
{original_task_description}

### Reversed Task Objective
Now, your task is to generate a reversed task objective based on the provided task description. The reversed objective should shift the focus from achieving the task goal to intentionally introducing noise into the data. Instead of performing actions such as classification, imputation, or any other task goal, the goal is to create challenges or distortions in the data. For example, if the original task involves classification, the reversed task should focus on introducing noise such as mislabeling or irrelevant features in the data.

### Data Examples
Here are the provided data examples related to the original task:

- {example_1}
- {example_2}
- {example_3}
- ...

### Generated Reversed Task Objective
{generated_reversed_task}
\end{lstlisting}

\renewcommand\lstlistingname{Prompt}

\begin{lstlisting}[breaklines=true,label={prompt:noise},caption={Prompt for noisy data synthesis.}]
### Reversed Task Objective
You are given the following reversed task objective. This objective describes how to intentionally introduce noise into the dataset.

{reversed_task_objective}

### Data Examples
Here are some sample data records that illustrate the structure and format of the dataset:

- {example_1}
- {example_2}
- {example_3}
- ...

### Instruction
Write executable Python code that introduces the noise into the dataset as described in the reversed task objective.  
The code should:
1. Take as input a dataset file (format consistent with the given examples).
2. Implement the noise generation specified in the reversed task objective.
3. Output the modified dataset to required file path in the same format as the input.
4. Ensure reproducibility (e.g., by setting a random seed if randomness is used).

### Expected Output
Provide only the Python code that implements the noise injection process.  
The code must be complete and runnable.
\end{lstlisting}

\renewcommand\lstlistingname{Prompt}

\begin{lstlisting}[breaklines=true,label={prompt:eval},caption={Prompt for evaluation scripts generation.}]
### Task Description
You are given a data governance task description:

{task_description}

### Data Samples
Here are some representative ground truth (expected) data samples:

{gt_samples}

Here are some representative processed data samples:

{processed_samples}

### Instruction
Write a Python evaluation script that compares the processed dataset against the ground truth dataset and outputs a quantitative score between 0 and 1, reflecting the model’s effectiveness in completing the task.

The evaluation should:
1. Load the ground truth and processed datasets from file paths provided as arguments.
2. Use evaluation metrics appropriate for the task category:
   - Filtering: F1 Score (balance of precision and recall in filtering unwanted entries).
   - Refinement: Accuracy (correctness of standardized or transformed data fields).
   - Imputation: Completion Rate / Imputation Accuracy (ability to correctly fill in missing values).
   - Deduplication & Consistency: Duplicate Reduction Rate or Consistency Score (removal of duplicates or ensuring consistent values).
   - Data Integration: Integration Accuracy (accuracy of merging heterogeneous datasets, resolving conflicts).
   - Classification & Labeling: Accuracy, Precision, Recall, F1 Score (standard classification metrics).
3. Output the evaluation result as a dictionary with the key `"eval_score"` and the corresponding score (float between 0 and 1).
4. Print the dictionary as the final output.

### Expected Output
Provide only the Python code for the evaluation script.  
The code should be complete and runnable, following this template structure:

```python
def evaluate(processed_path):
    expected_path = get_gt()
    expected = load_gt(expected_path)
    processed = load_processed(processed_path)

    # implement task-specific evaluation logic here ...

    result = {"eval_score": <score>}
    print(result)
\end{lstlisting}

\renewcommand\lstlistingname{Prompt}

To enhance reproducibility and review transparency, this appendix discloses several prompts used in our experiments (including intent identification, pipeline assembly, operator retrieval, and code debugging). We emphasize that these prompts only support a subset of ``minimum viable" functionality and are not sufficient on their own to constitute the full contract‑driven Planner–Executor–Evaluator framework described in the main paper.
\renewcommand\lstlistingname{Prompt}

Prompt~\ref{prompt:planner} present the detailed prompts for Planner.

\begin{lstlisting}[breaklines=true,label={prompt:planner},caption={Prompt for Intent Understanding.}]
[Role] You are an intent analysis robot. You need to identify the user's explicit intent from the conversation and analyze the user's data processing requirements based on the conversation content.
[Task]

You need to determine whether the user's current requirement is for a single operator or a complete pipeline, and set is_single_operator (true only if a single operator is required, otherwise false) and is_pipeline (true if pipeline processing is required, otherwise false) accordingly.
You need to summarize the user's processing requirements in detail based on the conversation history, and always provide a natural language response as the value of assistant_reply.
[Input Content] Conversation history: {history} Current user request: {target}
[Output Rules]
Reply only in the specified JSON format.
Do not output anything except JSON.
[Example]
{
"is_single_operator": false,
"is_pipeline": true,
"assistant_reply": "I will recommend a suitable data processing pipeline based on your needs.",
"reason": "The user explicitly requested a recommendation, wants to process data related to mathematics, and hopes to generate pseudo-answers.",
"purpose": "According to the conversation history, the user does not need a deduplication operator, hopes to generate pseudo-answers, and wants to keep the number of operators at 3."
}
\end{lstlisting}

\renewcommand\lstlistingname{Prompt}

Prompt~\ref{prompt:recommend} is for the agent in recommend Module.

\begin{lstlisting}[breaklines=true,label={prompt:recommend},caption={Prompt for the agent in recommend Module.}]
[ROLE]
You are a data governance workflow recommendation system. Based on the provided context, automatically select the appropriate operator nodes and assemble them into a complete data processing pipeline.

[INPUT]
You will receive the following information:
- Workflow requirements to be satisfied:
  {workflow_bg}
- Sample data information:
  {local_tool_for_sample}
- List of available operators:
  {operators}

[OUTPUT RULES]
1. Select suitable operator nodes from the available operators and assemble them into a complete processing pipeline. Output in the following JSON format:
   {"edges":[{"source":node0,"target":node1},{"source":node1,"target":node2}]}
2. Provide your reasoning for the selection in the following JSON format:
   {"reason": "Please explain your reasoning in detail here. For example: The pipeline includes multi-level data preprocessing and quality filtering, performing language filtering, format standardization, noise removal, privacy protection, length and structure optimization, and symbol and special character handling sequentially to ensure the text content is standardized, rich, and compliant."}
3. Verify that the constructed pipeline satisfies all requirements, especially {workflow_bg}.
4. Check the edges field to ensure all nodes are valid node fields from the available operators.
5. For each operator, specify the conditions under which it can continue execution, using the following format:
   "node1": {
     "Score": { "operator": ">", "value": 0.5 }
   }.
\end{lstlisting}

\renewcommand\lstlistingname{Prompt}

Prompt~\ref{prompt:exe} Prompt for the agent in op lib Module.

\begin{lstlisting}[breaklines=true,label={prompt:exe},caption={Prompt for the agent in op lib Module.}]
[ROLE]
You are an expert in data operator retrieval.

[TASK]
Based on the provided operator content {get_operator_content}, user requirement {target}, and operator names {op_name}, identify the top {top-k} most similar operator names from the operator library and provide your reasoning.

[INPUT FORMAT]
The input includes:
- Operator content (get_operator_content)
- User requirement (target)
- Operator names (op_name)

[OUTPUT RULES]
1. Strictly return the content in the JSON structure shown below. Do not include any extra content, comments, or additional fields.
2. You must return exactly {top-k} operator names in all cases.

JSON output example:
{
  "match_operators": [
    "OperatorName1",
    "OperatorName2",
    "OperatorName3",
    "OperatorName4"
  ],
  "reason": "xxx"
}
\end{lstlisting}

\renewcommand\lstlistingname{Prompt}

Prompt~\ref{prompt:execode} Prompt for the agent in write op Module.

\begin{lstlisting}[breaklines=true,label={prompt:execode},caption={Prompt for the agent in write op Module.}]
[ROLE]  
You are an expert in data operator development.

[TASK]  
Refer to the example operator {example} and write a new operator based on the requirements described in {target}.

[INPUT FORMAT]  
Input includes:  
- Example operator (example)  
- Target description (target)

[OUTPUT FORMAT]  
Please output in the following JSON structure:  
{  
  "code": "Complete source code of the operator",  
  "desc": "Brief description of the operator’s function and its input/output"  
}

[RULES]  
1. Carefully analyze and understand the structure and coding style of the example operator.  
2. Write operator code that fully meets the functional requirements of {target} and can run independently. Do not include any extra code or comments.  
3. Only output the two fields 'code' (the complete operator code as a string) and 'desc' (a concise explanation of the operator’s function and its input/output), strictly following the JSON format.  
4. If the operator requires using an LLM, the __init__ method must include the llm_serving field.  
5. All output files generated by the operator must be in the same directory as the current file (os.path.dirname(__file__)).
\end{lstlisting}

\renewcommand\lstlistingname{Prompt}

Prompt~\ref{prompt:debug} Prompt for the agent in debug Module.

\begin{lstlisting}[breaklines=true,label={prompt:debug},caption={Prompt for the agent in debug Module.}]
[ROLE]
You are an expert in code debugging and correction.

[TASK]
Given the original code, error message, requirement, JSON data fields, and reference code, minimally modify the original code to fix the error. Ensure your corrections are precise and focus on issues such as key alignment or import errors. Output the corrected code and your reason for modification strictly in JSON format, and follow all specified requirements.

[INPUT]
You will receive the following information:
- The original code: {code}
- The error message: {error}
- The requirement: {target}
- The JSON data fields processed in the target code: {data_keys}
- Reference code retrieved: {cls_detail_code}

[OUTPUT RULES]
1. Strictly return your response in JSON format, including: the complete corrected code, your reason for the modification, and any additional files that may be needed to better resolve the error. For example: {"code": xxx, "reason": xxx}
2. Ensure that the operator output file is in the same directory as the currently executing file (os.path.dirname(file)).
3. Do not include any extra keys, explanations, comments, or markdown syntax.
4. The returned code must include the if __name__ == '__main__': block, so that the file can be run independently.
5. The output must be in JSON format!!!!
6. You must use the files specified in <INPUT_FILES>{INPUT_FILES}</INPUT_FILES> as input.
\end{lstlisting}

\end{document}